%% file: paper.tex
\documentclass{new_tlp}
\usepackage{times,helvet,courier}
\usepackage{multirow}
\usepackage{latexsym}
\usepackage{amssymb}
\usepackage{xspace}
\usepackage{url} 
\usepackage{graphicx}
\usepackage{epsfig}
\usepackage{amsmath}
\usepackage[listofformat=parens,subrefformat=subparens,position=top]{subfig}
\usepackage{tikz}
\usetikzlibrary{arrows}
\usepackage{listings}
\let\origthelstnumber\thelstnumber
\makeatletter
\newcommand*\Suppressnumber{%
  \lst@AddToHook{OnNewLine}{%
    \let\thelstnumber\relax%
     \advance\c@lstnumber-\@ne\relax%
    }%
}
\newcommand*\Reactivatenumber{%
  \lst@AddToHook{OnNewLine}{%
   \let\thelstnumber\origthelstnumber%
   \advance\c@lstnumber\@ne\relax}%
}

\newtheorem{example}{Example}
\newcommand{\eofex}{\hfill$\blacksquare$}

\newcommand{\Probl}[1]{\textit{#1}}

\newif\ifdotikz\dotikzfalse

\ifdotikz
\usepackage{tikz}
\usetikzlibrary{positioning}
\pgfrealjobname{lpnmr13}
\usetikzlibrary{shapes,arrows,backgrounds,chains,%
matrix,patterns,arrows,decorations.pathmorphing,decorations.pathreplacing,%
positioning,fit,calc,decorations.text,shadows%
}
\else
\long\def\beginpgfgraphicnamed#1#2\endpgfgraphicnamed{\includegraphics{#1}}
\fi

\input{macros}

\title[The Seventh Answer Set Programming Competition: Design and Results]{The Seventh Answer Set Programming Competition: Design and Results%
}

  \author[M. Gebser, M. Maratea and F. Ricca]
         {
         	MARTIN GEBSER\\
         	Institute for Computer Science, University of Potsdam, Germany\\
         	\email{gebser@cs.uni-potsdam.de}
         	\and
         	MARCO MARATEA\\
         	DIBRIS, University of Genova, Italy\\
         	\email{marco@dibris.unige.it}
                \and 
                FRANCESCO RICCA\\
                Dipartimento di Matematica e Informatica, Universit{\`a} della Calabria, Italy\\
                \email{ricca@mat.unical.it}
		}

\jdate{March 2003}
\pubyear{2003}
\pagerange{\pageref{firstpage}--\pageref{lastpage}}
\doi{S1471068401001193}

\begin{document}
\label{firstpage}

\maketitle

\input{abstract}


\input{introduction}
\input{format}
 \input{benchmarks}
 \input{participants}

 \input{results}

\input{conclusions}

\paragraph{\bf Acknowledgments.} The organizers of the Seventh ASP Competition would like to thank the
LPNMR 2017 officials for the co-location of the event. We also
acknowledge the Department of Mathematics and Computer Science at the
University of Calabria for supplying the computational resources to
run the competition. Finally, we thank all solver and benchmark contributors, and participants, who worked hard to make this competition possible.

\input{paper.bbl}

\include{appendix}
\end{document}

%% file: macros.tex
\usepackage{framed}
\usepackage{amssymb}
\usepackage{latexsym}

\newcommand\nameFont[1]{{\fontencoding{OT1}\fontfamily{cmss}\selectfont{#1}}}
\def\nf{\nameFont}

\def\Mancoosi{{\sc Mancoosi}\xspace}

\def\aspcoretwo{\nf{ASP-Core-2}\xspace}

\newcommand\nop[1]{}

\newcommand{\negspace}{\vspace{-0.8mm}}

\newenvironment{indentnew}[1]%
{\begin{list}{}%
         {\setlength{\leftmargin}{#1}}%
         \item[]%
} {\end{list}}

\newcommand{\lesskip}{\hspace*{-0.5em}}
\newcommand{\hreduce}{\hspace*{-1.4mm}}
\newcommand{\gap}{\\[-3mm]}

\newcommand{\Dec}{D} 
\newcommand{\Opt}{O} 
\newcommand{\Que}{Q} 

\mathchardef\hn="2D

\def\clasp{{\sc {\small clasp}}} 

\def\Gringo{{\sc {\small gringo}}} 

\def\lptwoclaspold{{\sc {\small lp2normal2+clasp}}}
\def\waspdlvold{{\sc {\small wasp-1.5}}}

%% file: abstract.tex
\begin{abstract}
Answer Set Programming (ASP) is a prominent knowledge representation language with roots in logic programming and non-monotonic reasoning.
Biennial ASP competitions are organized in order to furnish challenging benchmark collections and assess the advancement of the state of the art in ASP solving.
In this paper, we report on the design and results of the Seventh ASP Competition, 
jointly organized by the University of Calabria (Italy), the University of Genova (Italy), and the University of Potsdam (Germany), 
in affiliation with the 14th International Conference on Logic Programming and Non-Monotonic Reasoning (LPNMR 2017). (Under consideration foracceptance in TPLP).
%
\end{abstract}

%% file: introduction.tex
\section{Introduction}\label{sec:introduction}

Answer Set Programming (ASP) is a prominent knowledge representation language with roots in logic programming and non-monotonic reasoning \cite{bara-2002,%
 DBLP:journals/cacm/BrewkaET11,%
 eite-etal-2009-primer,%
 gelf-leon-02,%
 lifs-2002,
 mare-trus-99,%
 niem-99}.
 The goal of the ASP Competition series is to promote advancements in ASP methods,
 collect challenging benchmarks, and assess the
 state of the art in ASP solving (see, e.g.,
 \cite{
 alvi-etal-2015-lpnmr,%
 alcadofuleperiveza17a,%
 brblbocapojalaradeve15a,%
 gekakarosc15a,%
 lebestga17a,%
 mara-etal-2015-lpnmr,%
 margup14a,%
DBLP:journals/ia/CalimeriFPZ17} for recent ASP systems, and \cite{gebs-etal-2018-ijcai} for a recent survey). 
Following a nowadays customary practice of publishing results of AI-based competitions in archival journals, where they are expected to remain available and can be used as references, the results of ASP competitions have been hosted in prominent journals of the area (see,~\cite{CalimeriIR14,cali-etal-aij-2016,GebserMR17}).
Continuing the tradition, this paper reports on the design and results of the Seventh ASP Competition,\footnote{\url{http://aspcomp2017.dibris.unige.it}} which was jointly organized by the University of Calabria (Italy), the University of Genova (Italy), and the University of Potsdam (Germany), in affiliation with the 14th International Conference on Logic Programming and Non-Monotonic Reasoning (LPNMR 2017).\footnote{\url{http://lpnmr2017.aalto.fi}} 

The Seventh ASP Competition is conceived along the lines of the System track of previous competition editions \cite{cali-etal-aij-2016,lier-etal-aima-2016,gebs-etal-aaai-2016,GebserMR17}, with the following characteristics:
$(i)$ benchmarks adhere to the \aspcoretwo standard modeling language,%
\footnote{\url{http://www.mat.unical.it/aspcomp2013/ASPStandardization/}}
$(ii)$ sub-tracks are based on language features utilized in problem encodings
(e.g., aggregates, choice or disjunctive rules, queries, and weak constraints), and
$(iii)$ problem instances are classified and selected according to their expected hardness. 
%
Both single and multi-processor categories are available in the competition, where solvers in the first category 
run on a single CPU (core), while they can take advantage of multiple processors (cores) in the second category. 
In addition to the basic competition design,
which has also been addressed in a preliminary version of this report \cite{gemari17a},
we detail the revised benchmark selection process as well as the results of the event,
which were orally presented during LPNMR 2017 in Hanasaari, Espoo, Finland.

The rest of this paper is organized as follows.
Section~\ref{sec:format} introduces the format of the Seventh ASP Competition.
In Section~\ref{sec:benchmarks},
we describe new problem domains contributed to this competition edition as well as
the revised benchmark selection process for picking instances to run in the competition.
The participant systems of the competition are then surveyed in Section~\ref{sec:participants}.
In Section~\ref{sec:results}, we then present the results of the Seventh ASP Competition along
with the winning systems of competition categories. 
Section~\ref{sec:conclusions} concludes the paper with final remarks.


%% file: format.tex
\section{Competition Format}
\label{sec:format}
\newsavebox{\tspinstance}
\begin{lrbox}{\tspinstance}
\begin{minipage}{60.5mm}
\begin{lstlisting}
node(1). edge(1,2). cost(1,2,3). 
         edge(1,4). cost(1,4,1).§\vspace{-2mm}\addtocounter{lstnumber}{-1}§

node(2). edge(2,1). cost(2,1,2).
         edge(2,3). cost(2,3,1).§\vspace{-2mm}\addtocounter{lstnumber}{-1}§

node(3). edge(3,2). cost(3,2,2).
         edge(3,4). cost(3,4,2).§\vspace{-2mm}\addtocounter{lstnumber}{-1}§

node(4). edge(4,1). cost(4,1,2).
         edge(4,3). cost(4,3,2).
\end{lstlisting}
\end{minipage}
\end{lrbox}
\newsavebox{\tspbasic}
\begin{lrbox}{\tspbasic}
\begin{minipage}{111.5mm}
\begin{lstlisting}
cycle(X,Y) :- edge(X,Y), edge(X,Z), Y != Z, not cycle(X,Z).§\vspace{-2mm}\addtocounter{lstnumber}{-1}§

reach(Y) :- cycle(1,Y).
reach(Y) :- cycle(X,Y), reach(X).§\vspace{-2mm}\addtocounter{lstnumber}{-1}§

:- node(Y), not reach(Y).
\end{lstlisting}
\end{minipage}
\end{lrbox}
\newsavebox{\tspadvanced}
\begin{lrbox}{\tspadvanced}
\begin{minipage}{111.5mm}
\begin{lstlisting}
{cycle(X,Y) : edge(X,Y)} = 1 :- node(X).§\vspace{-2mm}\addtocounter{lstnumber}{-1}§

reach(Y) :- cycle(1,Y).
reach(Y) :- cycle(X,Y), reach(X).§\vspace{-2mm}\addtocounter{lstnumber}{-1}§

:- node(Y), not reach(Y).
\end{lstlisting}
\end{minipage}
\end{lrbox}
\newsavebox{\tspdisjunctive}
\begin{lrbox}{\tspdisjunctive}
\begin{minipage}{111.5mm}
\begin{lstlisting}
cycle(1,Y) | cycle(1,Z) :- edge(1,Y), edge(1,Z), Y != Z.
cycle(X,Y) | cycle(X,Z) :- edge(X,Y), edge(X,Z), Y != Z,§\Suppressnumber§
                                       reach(X), X != 1.§\vspace{-2mm}\addtocounter{lstnumber}{-1}\Reactivatenumber§

reach(Y) :- cycle(X,Y).§\vspace{-2mm}\addtocounter{lstnumber}{-1}§

:- node(Y), not reach(Y).
\end{lstlisting}
\end{minipage}
\end{lrbox}
\newsavebox{\tspoptimize}
\begin{lrbox}{\tspoptimize}
\begin{minipage}{111.5mm}
\begin{lstlisting}
:§$\scriptstyle{\sim}\hspace{-0.5pt}$§ cycle(X,Y), cost(X,Y,C). [C,X,Y]
\end{lstlisting}
\end{minipage}
\end{lrbox}
\begin{figure}[t]
\subfloat[A directed graph with edge costs.\label{fig:tsp:graph}]{%
\begin{tikzpicture}[->,>=stealth',auto,node distance=2.5cm,inner sep=2.5pt,thick]
\node[circle,draw, minimum height=5mm, inner sep=0pt] (A)              {1};
\node[circle,draw, minimum height=5mm, inner sep=0pt] (B) [right of=A] {2};
\node[circle,draw, minimum height=5mm, inner sep=0pt] (C) [below of=B] {3};
\node[circle,draw, minimum height=5mm, inner sep=0pt] (D) [below of=A] {4};
\path (A) edge [bend left] node {$3$} (B);
\path (B) edge [bend left] node {$2$} (A);
\path (B) edge [bend left] node {$1$} (C);
\path (C) edge [bend left] node {$2$} (B);
\path (C) edge [bend left] node {$2$} (D);
\path (D) edge [bend left] node {$2$} (C);
\path (A) edge [bend left] node {$1$} (D);
\path (D) edge [bend left] node {$2$} (A);
\end{tikzpicture}}%
\hspace{10.4mm}
\subfloat[Fact representation of the graph in \protect\subref*{fig:tsp:graph}.\label{fig:tsp:data}]{
\usebox{\tspinstance}}%
\\
\subfloat[Basic Decision encoding of Hamiltonian cycles.\label{fig:tsp:basic}]{
\usebox{\tspbasic}}%
\\
\subfloat[Advanced Decision encoding of Hamiltonian cycles.\label{fig:tsp:advanced}]{
\usebox{\tspadvanced}}%
\\
\subfloat[Unrestricted encoding of Hamiltonian cycles.\label{fig:tsp:disjunctive}]{
\usebox{\tspdisjunctive}}%
\\
\subfloat[Weak constraint for Hamiltonian cycle optimization.\label{fig:tsp:optimize}]{
\usebox{\tspoptimize}}%
\caption{An example graph with edge costs, its fact representation, and corresponding encodings.\label{fig:tsp}}%
\end{figure}
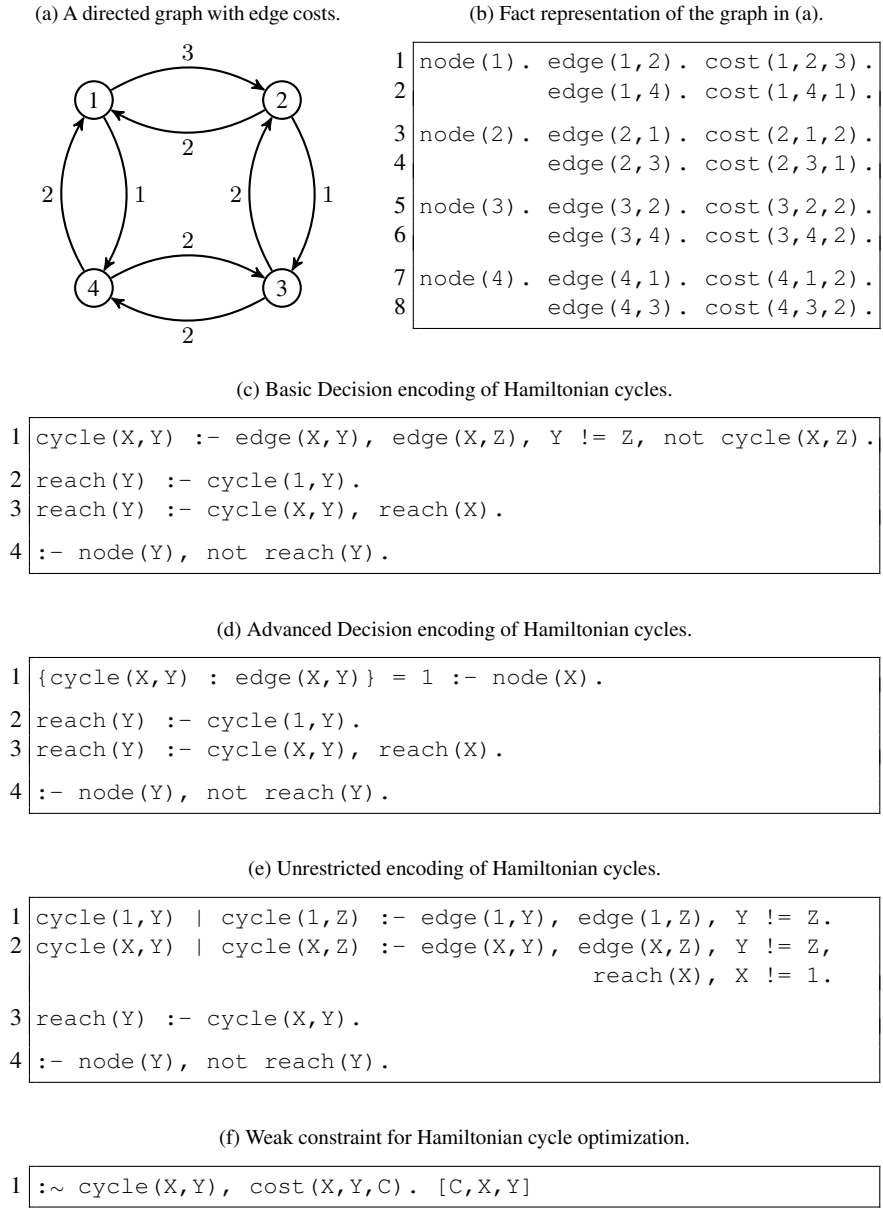

This section gives an overview of competition categories, sub-tracks, and scoring scheme(s),
which are similar to the previous ASP Competition edition.
One addition though concerns the system ranking of Optimization problems,
where a ranking by the number of instances solved ``optimally''
complements the relative scoring scheme based on solution quality used previously.

\paragraph{Categories.} The competition includes two categories, depending on the computational resources provided to participant systems:
    {\bf SP}, where one processor (core) is available, and 
    {\bf MP}, where multiple processors (cores) can be utilized.
While the {\bf SP} category aims at sequential solving systems,
{\bf MP} allows for exploiting parallelism. 


\paragraph{Sub-tracks.} 
Both categories 
are structured into the following four sub-tracks,
based on the \aspcoretwo\ language features utilized in problem encodings:
%
\begin{itemize}
\item[$\bullet$]
    \textbf{Sub-track \#1} ({\em Basic Decision}):
    Encodings consisting of non-disjunctive and non-choice rules (also called normal rules) with classical and built-in atoms only.
\item[$\bullet$]
    \textbf{Sub-track \#2} ({\em Advanced Decision}):
    Encodings exploiting the language fragment allowing for aggregates, choice as well as disjunctive rules, and queries,
    yet excepting weak constraints and non-head-cycle-free (non-HCF) disjunction.
\item[$\bullet$]
    \textbf{Sub-track \#3} ({\em Optimization}):
    Encodings extending the aforementioned language fragment by weak constraints, while still excepting non-HCF disjunction.
\item[$\bullet$]
    \textbf{Sub-track \#4} ({\em Unrestricted}):
    Encodings exploiting the full language and, in particular, non-HCF disjunction.
\end{itemize}
A problem domain, i.e., an encoding together with a collection of instances,
belongs to the first sub-track its problem encoding is compatible with.

\begin{example}\label{ex:tsp}
To illustrate the sub-tracks and respective language features,
consider the directed graph displayed in Figure~\subref{fig:tsp:graph}
and the corresponding fact representation given in Figure~\subref{fig:tsp:data}.
Facts over the predicate \lstinline{node}/1 specify the nodes of the graph,
those over \lstinline{edge}/2 provide the edges, and \lstinline{cost}/3
associates each edge with its cost.
The idea in the following is to encode the well-known Traveling Salesperson problem,
which is about finding a Hamiltonian cycle, i.e., a round trip visiting each node exactly once,
such that the sum of edge costs is minimal.
Note that the example graph in Figure~\subref{fig:tsp:graph} includes precisely two outgoing
edges per node, and for simplicity the encodings in Figures~\subref{fig:tsp:basic}--\subref*{fig:tsp:disjunctive} build on this property,
while accommodating an arbitrary number of outgoing edges would also be possible with appropriate modifications.

The first encoding in Figure~\subref{fig:tsp:basic} complies with the language fragment
of Sub-track \#1, as it does not make use of aggregates, choice or disjunctive rules,
queries, and weak constraints.
Note that terms starting with an uppercase letter,
such as \lstinline{X}, \lstinline{Y}, and \lstinline{Z},
stand for universally quantified first-order variables,
\lstinline{Y != Z} is a built-in atom, and
\lstinline{not} denotes the (default) negation connective.
Given this,
the rule in line~1 expresses that exactly one of the two outgoing edges
per node must belong to a Hamiltonian cycle,
represented by atoms over the predicate \lstinline{cycle}/2 within a stable model
\cite{lifschitz08a}.
Starting from the distinguished node~\lstinline{1},
the least fixpoint of the rules in lines~2 and~3 provides the nodes
reachable from~\lstinline{1} via the edges of a putative Hamiltonian cycle.
The so-called integrity constraint, i.e., a rule with an empty head that is
interpreted as false, in line~4 then asserts that all nodes must be reachable
from the starting node~\lstinline{1},
which guarantees that stable models coincide with Hamiltonian cycles.
While edge costs are not considered so far,
the encoding in Figure~\subref{fig:tsp:basic} can be used to decide whether
a Hamiltonian cycle exists for a given graph (with precisely two outgoing
edges per node).

The second encoding in Figure~\subref{fig:tsp:advanced} includes a choice
rule in line~1, thus making use of language features permitted in Sub-track \#2,
but incompatible with Sub-track \#1.
The instance of this choice rule obtained for the node~\lstinline{1},
\lstinline|{cycle(1,2); cycle(1,4)} = 1.|,
again expresses that exactly one outgoing edge of node~\lstinline{1}
must be included in a Hamiltonian cycle,
and respective rule instances apply to the other nodes of the example graph
in Figure~\subref{fig:tsp:graph}.
Notably, the choice rule adapts to an arbitrary number of outgoing edges,
and the assumption that there are precisely two per node could be dropped
when using the encoding in Figure~\subref{fig:tsp:advanced}.

The rules in lines~1 and~2 of the third encoding in Figure~\subref{fig:tsp:disjunctive}
are disjunctive, and rule instances as follows are obtained together with line~3:

\begin{lstlisting}[frame=none,numbers=none]

cycle(1,2) | cycle(1,4).
cycle(2,1) | cycle(2,3) :- reach(2).
cycle(3,2) | cycle(3,4) :- reach(3).
cycle(4,1) | cycle(4,3) :- reach(4).
reach(1) :- cycle(2,1).    reach(3) :- cycle(2,3). 
reach(1) :- cycle(4,1).    reach(3) :- cycle(4,3).
reach(2) :- cycle(1,2).    reach(2) :- cycle(3,2).
reach(4) :- cycle(1,4).    reach(4) :- cycle(3,4).

\end{lstlisting}

Observe that \lstinline{reach(3)} occurs in the body of a disjunctive
rule with \lstinline{cycle(3,2)} and \lstinline{cycle(3,4)} in the head.
These atoms further imply \lstinline{reach(2)} or \lstinline{reach(4)},
respectively, which lead on two disjunctive rules,
one containing \lstinline{cycle(2,3)} in the head and the other \lstinline{cycle(4,3)}.
As the latter two atoms also occur in the body of rules with \lstinline{reach(3)}
in the head, we have that all of the mentioned atoms recursively depend on each other.
Since \lstinline{cycle(3,2)} and \lstinline{cycle(3,4)} jointly constitute the head of
a disjunctive rule, this means that rule instances obtained from the encoding in
Figure~\subref{fig:tsp:disjunctive} are non-HCF \cite{bene-dech-94} and thus fall into
a syntactic class of logic programs able to express problems
at the second level of the polynomial hierarchy \cite{EiterG95}.
Hence, the encoding in Figure~\subref{fig:tsp:disjunctive} makes use of a language feature
permitted in Sub-track \#4 only.

Given that either of the encodings in Figures~\subref{fig:tsp:basic}--\subref*{fig:tsp:disjunctive}
yields stable models corresponding to Hamiltonian cycles,
the weak constraint in Figure~\subref{fig:tsp:optimize} can be added to each of them
to express the objective of finding a Hamiltonian cycle whose sum of edge costs is minimal.
In case of the encodings in Figures~\subref{fig:tsp:basic} and~\subref{fig:tsp:advanced},
the addition of the weak constraint leads to a reclassification into Sub-track \#3,
since the focus is shifted from a Decision to an Optimization problem.
For the encoding in Figure~\subref{fig:tsp:disjunctive}, Sub-track \#4 still matches  
when adding the weak constraint, as non-HCF disjunction is excluded in the other sub-tracks.%
\eofex
\end{example}

\paragraph{Scoring Scheme.} 
The applied scoring schemes are based on the following considerations:
\begin{itemize}
\item
    All domains are weighted equally. 
\item
    If a system outputs an incorrect answer to some instance in a domain,
    this invalidates its score for the domain,
    even if other instances are solved correctly.
\end{itemize}
In general, $100$ points can be earned in each problem domain.
The total score of a system is the sum of points over all
domains.

For {\em Decision} problems and {\em Query answering} tasks, the score $S(D)$ of a system~$S$ in a domain~$D$
featuring $N$ instances is calculated as
\begin{eqnarray*}
S(D) & = & \frac{N_S * 100}{N}
\end{eqnarray*}
where $N_S$ is the number of instances successfully solved within the time and memory limits
of 20 minutes wall-clock time and 12GB RAM per run.

For {\em Optimization} problems, we employ two alternative scoring schemes.
The first one, which has also been used in the previous competition edition,
performs a relative ranking of systems by solution quality,
following the approach of the \Mancoosi International Solver Competition.%
\footnote{\url{http://www.mancoosi.org/misc/}}
Given $M$ 
participant systems
, the score $S(D,I)$ of a system~$S$ for an instance~$I$ in a domain~$D$ featuring $N$ instances is calculated as
\begin{eqnarray*}
    S(D,I) & = & \frac{M_S(I) * 100}{M * N}
\end{eqnarray*}
where $M_S(I)$ is
\begin{itemize}
\item
    $0$, if $S$ did neither produce a solution nor report unsatisfiability; or otherwise
\item
    the number of participant systems that did not produce any strictly better solution than $S$, where a confirmed optimum solution is considered strictly better than an unconfirmed one.
\end{itemize}
The score $S_1(D)$ of system~$S$ in domain~$D$ is then taken as the sum of scores $S(D,I)$ over the $N$ instances~$I$ in~$D$. 

The second scoring scheme considers the number of instances solved ``optimally'', i.e.,
a confirmed optimum solution or unsatisfiability is reported.
Hence, the score $S_2(D)$ of a system $S$ in a domain $D$ is defined as $S(D)$ above,
with $N_S$ being the number of instances solved optimally. 
This second scoring scheme (inspired by the MaxSAT Competition) gives more importance to solvers that can actually solve instances to the optimum, but it does not consider ``non-optimal'' solutions. The two measures provide alternative perspectives on the performance of participants solving optimization problems.

Note that, as with Decision problems and Query answering tasks,
$S_1(D)$ and $S_2(D)$ range from $0$ to $100$ in each domain.
$S_1(D)$ focuses on the best solutions found by participant systems, while
$S_2(D)$ on completed runs.

In each category and respective sub-tracks, the participant systems are ranked by their sums of scores over all domains, in decreasing order. In case of a draw in terms of the sum of scores, the sums of runtimes over all instances are taken into account as a tie-breaking criterion.


%% file: benchmarks.tex
\section{Benchmark Suite and Selection}\label{sec:benchmarks}

The benchmark suite of the Seventh ASP Competition includes 36 domains,
where 28 stem from the previous competition edition \cite{GebserMR17}, and
8 domains, as well as additional instances for the \Probl{Graph Colouring} problem,
were newly submitted.
We first describe the eight new domains and then detail the
instance selection process based on empirical hardness.


\subsection{New Domains}\label{sec:benchmarks:domains}

The eight new domains of this ASP Competition edition can be roughly characterized
as closely related to machine learning
(\Probl{Bayesian Network Learning},
 \Probl{Markov Network Learning}, and
 \Probl{Supertree Construction}),
personnel scheduling
(\Probl{Crew Allocation} and
\Probl{Resource Allocation}), or
combinatorial problem solving
(\Probl{Paracoherent Answer Sets},
 \Probl{Random Disjunctive ASP}, and
 \Probl{Traveling Salesperson}), respectively.
While \Probl{Traveling Salesperson} constitutes a classical optimization problem
in computer science, the five domains stemming from machine learning and
personnel scheduling are application-oriented, and the contribution of such practically
relevant benchmarks to the ASP Competition is particularly encouraged
\cite{GebserMR17}.
Moreover, the \Probl{Paracoherent Answer Sets} and \Probl{Random Disjunctive ASP}
domains contribute to Sub-track \#4,
which was sparsely populated in recent ASP Competition editions,
and beyond theoretical interest these benchmarks are relevant
to logic program debugging and industrial solvers development.
The following paragraphs provide more detailed background information
for each of the new domains. 

\paragraph{\Probl{Bayesian Network Learning}.}
Bayesian networks are directed acyclic graphs representing
(in)dependence relations between variables in multivariate data analysis.
Learning the structure of Bayesian networks, i.e., selecting arcs such that the
resulting graph fits given data best, is a combinatorial optimization problem
amenable to constraint-based solving methods like the one proposed in \cite{Cussens11:uai}.
In fact, data sets from the literature serve as instances in this domain,
while a problem encoding in \aspcoretwo expresses optimal Bayesian networks,
given by directed acyclic graphs whose associated cost is minimal.

\paragraph{\Probl{Crew Allocation}.}
This scheduling problem,
which has also been addressed by related constraint-based solving methods \cite{guecan95a},
deals with allocating crew members to flights
such that the amount of personnel with certain capabilities
(e.g., role on board and spoken language) as well as
off-times between flights are sufficient.
Moreover, instances with different numbers of flights and available personnel
restrict the amount of personnel that may be allocated to flights
in such a way that no schedule is feasible under the given restrictions.

\paragraph{\Probl{Markov Network Learning}.}
As with Bayesian networks, the learning problem for Markov networks \cite{jagerinypeco15a}
aims at the optimization of graphs representing the dependence structure between variables
in statistical inference.
In this domain, the graphs of interest are undirected and required to be chordal,
while associated scores express marginal likelihood with respect to given data.
Problem instances of varying hardness are obtained by taking samples of
different size and density from literature data sets.

\paragraph{\Probl{Resource Allocation}.}
This scheduling problem deals with
allocating the activities of business processes to human resources such that
role requirements and temporal relations between activities are met \cite{hacamepo16a}.
Moreover, the total makespan of schedules is subject to an upper bound
as well as optimization.
The hardness of instances in this domain varies with respect to
the number of activities, temporal relations, available resources, and upper bounds.

\paragraph{\Probl{Supertree Construction}.}
The goal of the supertree construction problem \cite{kooijasa15a}
is to combine the leaves of several given phylogenetic subtrees
into a single tree fitting the given subtrees as closely as possible.
That is, optimization aims at preserving the structure of subtrees, 
where the introduction of intermediate nodes between direct neighbors is tolerated,
while the avoidance of such intermediate nodes is an optimization target as well.
Instances of varying hardness are obtained by mutating projections
of binary trees with different numbers of leaves. 

\paragraph{\Probl{Traveling Salesperson}.}
The well-known traveling salesperson problem \cite{apbichco07a}
is to find a round trip through a (directed) graph that is optimal in terms of the
accumulated edge costs.
Instances in this domain are twofold by
stemming from the TSPLIB repository%
\footnote{\url{http://elib.zib.de/pub/mp-testdata/tsp/tsplib/tsplib.html}}
or being randomly generated to increase the variety in the ASP Competition, respectively.

\paragraph{\Probl{Paracoherent Answer Sets}.}
Given an incoherent logic program~$P$, i.e., a program~$P$ without answer sets,
a paracoherent (or semi-stable) answer set corresponds to a gap-minimal answer set of the epistemic transformation of~$P$ \cite{inou-saka-95,DBLP:journals/ai/AmendolaEFLM16}.
The instances in this domain,
used in \cite{DBLP:conf/aaai/AmendolaDFLR17,DBLP:conf/aaai/AmendolaDFR18} to evaluate genuine implementations of paracoherent ASP,
are obtained by grounding and transforming incoherent programs 
from previous editions of the ASP Competition.
In particular,
weak constraints single out answer sets of
a transformed program containing a minimal number of atoms that are
actually underivable from the original program.

\paragraph{\Probl{Random Disjunctive ASP}.}
The disjunctive logic programs in this domain
\cite{DBLP:conf/ijcai/AmendolaRT17} express random 2QBF formulas,
given as conjunctions of terms in disjunctive normal form,
by an extension of the Eiter-Gottlob encoding    \cite{EiterG95}.
Parameters controlling the random generation of 2QBF formulas
(e.g., number of variables and number of conjunctions)
are  set so that instances lie close to the phase transition region,
while having an expected average solving time below the competition timeout of 20 minutes per run.
%

\subsection{Benchmark Selection}\label{sec:benchmarks:selection}

\begin{table}
\centering
\hreduce
  \begin{tabular}{|@{\,}l@{\,}|@{\!\!\!}c@{\,}|@{\!\!\!}r@{\!\!\!\!}r@{\!}r@{\!\!\!\!}r@{\,}|@{\!\!\!}r@{\!\!\!\!}r@{\!}r@{\!\!\!\!}r@{\,}|@{\!\!\!}r@{\!\!\!\!}r@{\!}r@{\!\!\!\!}r@{\,}|@{\!\!\!}r@{\!\!\!\!}r@{\!}r@{\!\!\!\!}r@{\,}|l|}
  \cline{1-18}
  \textbf{Domain} & \nop{\textbf{App} &} \textbf{P} & 
  \multicolumn{4}{@{\!\!}c|@{\!\!\!}}{\,\,\textbf{Easy}} &
  \multicolumn{4}{@{\!\!\!}c|@{\!\!\!}}{\,\,\textbf{Medium}} &
  \multicolumn{4}{@{\!\!\!}c|@{\!\!\!}}{\,\,\textbf{Hard}} &
  \multicolumn{4}{@{\!\!\!}c|}{\,\,\textbf{Too hard}}
  \\\cline{1-18}\multicolumn{3}{c}{}\gap\cline{1-19}
  \Probl{Graph Colouring} & \nop{&} \Dec & 
  1 & (1) & 3 & (5) & 2 & (2) & 4 & (21) & 2 & (3) & 5 & (16) & 0 & (0) & 3 & (3) &                                               
  \multirow{5}{*}{\!\!\!\!\rotatebox[origin=c]{270}{\textbf{Sub-track \#1}}}
  \\\cline{1-18}
  \Probl{Knight Tour with Holes} & \nop{&} \Dec & 
  2 & (5) & 3 & (4) & 4 & (4) & 0 & (0) & 4 & (9) & 0 & (0) & 0 & (0) & 7 & (302) &
  \\\cline{1-18}
  \Probl{Labyrinth} & \nop{&} \Dec & 
  4 & (45) & 0 & (0) & 5 & (72) & 0 & (0) & 7 & (83) & 0 & (0) & 0 & (0) & 4 & (8) &
  \\\cline{1-18}
  \Probl{Stable Marriage} & \nop{&} \Dec & 
  0 & (0) & 0 & (0) & 3 & (3) & 0 & (0) & 6 & (15) & 1 & (1) & 0 & (0) & 10 & (55) &
  \\\cline{1-18}
  \Probl{Visit-all} & \nop{&} \Dec & 
  8 & (14) & 0 & (0) & 5 & (5) & 0 & (0) & 7 & (40) & 0 & (0) & 0 & (0) & 0 & (0) &
  \\\cline{1-19}\multicolumn{3}{c}{}\gap\cline{1-19} 
  \Probl{Combined Configuration} & \nop{$\surd$ &} \Dec & 
  1 & (1) & 0 & (0) & 1 & (1) & 0 & (0) & 12 & (44) & 0 & (0) & 0 & (0) & 6 & (34) &
  \multirow{12}{*}{\!\!\!\!\rotatebox[origin=c]{270}{\textbf{Sub-track \#2}}}
  \\\cline{1-18}
  \Probl{Consistent Query Answering} & \nop{$\surd$ &} \Que & 
  0 & (0) & 0 & (0) & 0 & (0) & 0 & (0) & 0 & (0) & 0 & (0) & 0 & (0) & 20 & (120) &
  \\\cline{1-18}
  \textbf{\Probl{Crew Allocation}} & \nop{$\surd$ &} \Dec & 
  0 & (0) & 4 & (10) & 0 & (0) & 6 & (11) & 0 & (0) & 6 & (10) & 0 & (0) & 4 & (6) &
  \\\cline{1-18}    
  \Probl{Graceful Graphs} & \nop{&} \Dec & 
  3 & (3) & 0 & (0) & 4 & (4) & 1 & (1) & 4 & (28) & 2 & (2) & 0 & (0) & 6 & (21) &
  \\\cline{1-18}
  \Probl{Incremental Scheduling} & \nop{$\surd$ &} \Dec & 
  2 & (11) & 2 & (6) & 3 & (47) & 2 & (11) & 3 & (37) & 2 & (10) & 0 & (0) & 6 & (76) &
  \\\cline{1-18}
  \Probl{Nomystery} & \nop{&} \Dec & 
  4 & (4) & 0 & (0) & 4 & (5) & 0 & (0) & 4 & (10) & 0 & (0) & 0 & (0) & 8 & (32) &
  \\\cline{1-18}
  \Probl{Partner Units} & \nop{$\surd$ &} \Dec & 
  3 & (9) & 1 & (1) & 4 & (34) & 0 & (0) & 3 & (15) & 1 & (1) & 0 & (0) & 8 & (32) &
  \\\cline{1-18}
  \Probl{Permutation Pattern Matching} & \nop{&} \Dec & 
  2 & (16) & 2 & (32) & 2 & (14) & 2 & (58) & 0 & (0) & 5 & (14) & 0 & (0) & 7 & (20) &
  \\\cline{1-18}
  \Probl{Qualitative Spatial Reasoning} & \nop{&} \Dec & 
  5 & (35) & 4 & (35) & 4 & (34) & 2 & (19) & 3 & (7) & 2 & (2) & 0 & (0) & 0 & (0) &
  \\\cline{1-18}
  \Probl{Reachability} & \nop{&} \Que & 
  0 & (0) & 0 & (0) & 10 & (30) & 10 & (30) & 0 & (0) & 0 & (0) & 0 & (0) & 0 & (0) &
  \\\cline{1-18}
  \Probl{Ricochet Robots} & \nop{&} \Dec & 
  2 & (2) & 0 & (0) & 7 & (18) & 0 & (0) & 4 & (181) & 0 & (0) & 0 & (0) & 7 & (38) &
  \\\cline{1-18}
  \Probl{Sokoban} & \nop{&} \Dec & 
  2 & (77) & 2 & (10) & 2 & (84) & 2 & (8) & 5 & (114) & 2 & (12) & 0 & (0) & 5 & (620) &
  \\\cline{1-19}\multicolumn{3}{c}{}\gap\cline{1-19} 
  \textbf{\Probl{Bayesian Network Learning}} & \nop{&} \Opt & 
  4 & (4) & 0 & (0) & 4 & (8) & 0 & (0) & 8 & (19) & 0 & (0) & 4 & (20) & 0 & (6) &
  \multirow{13}{*}{\!\!\!\!\rotatebox[origin=c]{270}{\textbf{Sub-track \#3}}}
  \\\cline{1-18}  
  \Probl{Connected Still Life} & \nop{&} \Opt & 
  0 & (0) & 0 & (0) & 5 & (5) & 0 & (0) & 10 & (70) & 0 & (0) & 5 & (45) & 0 & (0) &
  \\\cline{1-18}
  \Probl{Crossing Minimization} & \nop{$\surd$ &} \Opt & 
  1 & (1) & 0 & (0) & 1 & (1) & 0 & (0) & 17 & (80) & 0 & (0) & 1 & (1) & 0 & (0) &
  \\\cline{1-18}
  \textbf{\Probl{Markov Network Learning}} & \nop{&} \Opt & 
  0 & (0) & 0 & (0) & 0 & (0) & 0 & (0) & 0 & (0) & 0 & (0) & 10 & (10) & 10 & (50) &
  \\\cline{1-18}  
  \Probl{Maximal Clique} & \nop{&} \Opt & 
  0 & (0) & 0 & (0) & 0 & (0) & 0 & (0) & 10 & (41) & 0 & (0) & 10 & (94) & 0 & (1) &
  \\\cline{1-18}
  \Probl{MaxSAT} & \nop{$\surd$ &} \Opt & 
  0 & (0) & 0 & (0) & 4 & (4) & 0 & (0) & 0 & (0) & 0 & (0) & 0 & (0) & 16 & (50) &
  \\\cline{1-18}
  \textbf{\Probl{Resource Allocation}} & \nop{$\surd$ &} \Opt & 
  -- & (3) & -- & (0) & -- & (3) & -- & (0) & -- & (0) & -- & (0) & -- & (0) & -- & (0) &
  \\\cline{1-18}  
  \Probl{Steiner Tree} & \nop{$\surd$ &} \Opt & 
  0 & (0) & 0 & (0) & 0 & (0) & 0 & (0) & 1 & (1) & 0 & (0) & 16 & (45) & 3 & (3) &
  \\\cline{1-18}
  \textbf{\Probl{Supertree Construction}} & \nop{&} \Opt & 
  0 & (0) & 0 & (0) & 0 & (0) & 0 & (0) & 6 & (30) & 0 & (0) & 14 & (30) & 0 & (0) &
  \\\cline{1-18}  
  \Probl{System Synthesis} & \nop{$\surd$ &} \Opt & 
  0 & (0) & 0 & (0) & 0 & (0) & 0 & (0) & 8 & (16) & 0 & (0) & 8 & (80) & 4 & (4) &
  \\\cline{1-18}
  \textbf{\Probl{Traveling Salesperson}} & \nop{&} \Opt & 
  0 & (0) & 0 & (0) & 2 & (2) & 0 & (0) & 3 & (3) & 0 & (0) & 12 & (60) & 3 & (3) &
  \\\cline{1-18}
  \Probl{Valves Location} & \nop{$\surd$ &} \Opt & 
  6 & (10) & 0 & (0) & 2 & (2) & 0 & (0) & 7 & (29) & 0 & (0) & 5 & (244) & 0 & (23) &
  \\\cline{1-18}
  \Probl{Video Streaming} & \nop{$\surd$ &} \Opt & 
  11 & (16) & 0 & (0) & 0 & (0) & 0 & (0) & 0 & (0) & 0 & (0) & 8 & (22) & 1 & (1) &
  \\\cline{1-19}\multicolumn{3}{c}{}\gap\cline{1-19} 
  \Probl{Abstract Dialectical Frameworks} & \nop{&} \Opt & 
  4 & (18) & 0 & (0) & 8 & (20) & 0 & (0) & 6 & (122) & 0 & (0) & 2 & (2) & 0 & (0) &
  \multirow{6}{*}{\!\!\!\!\rotatebox[origin=c]{270}{\textbf{Sub-track \#4}}}
  \\\cline{1-18}
  \Probl{Complex Optimization} & \nop{$\surd$ &} \Dec & 
  0 & (0) & 0 & (0) & 0 & (0) & 0 & (0) & 20 & (78) & 0 & (0) & 0 & (0) & 0 & (0) &
  \\\cline{1-18}
  \Probl{Minimal Diagnosis} & \nop{$\surd$ &} \Dec & 
  7 & (158) & 2 & (55) & 3 & (9) & 2 & (8) & 4 & (4) & 1 & (1) & 0 & (0) & 0 & (0) &
  \\\cline{1-18}
  \textbf{\Probl{Paracoherent Answer Sets}} & \nop{&} \Opt & 
  0 & (0) & 0 & (0) & 1 & (1) & 0 & (0) & 12 & (112) & 0 & (0) & 0 & (0) & 7 & (43) &
  \\\cline{1-18}
  \textbf{\Probl{Random Disjunctive ASP}} & \nop{&} \Dec & 
  0 & (0) & 0 & (0) & 0 & (0) & 0 & (0) & 5 & (48) & 13 & (73) & 0 & (0) & 2 & (2) &
  \\\cline{1-18}
  \Probl{Strategic Companies} & \nop{&} \Que & 
  0 & (0) & 0 & (0) & 0 & (0) & 0 & (0) & 0 & (0) & 0 & (0) & 0 & (0) & 20 & (37) &
  \\\cline{1-19} 
  \end{tabular}
  \caption{Problem domains of benchmarks for the Seventh ASP Competition, where
           entries in the \textup{\textbf{P}} column indicate \emph{Decision} (``\textup{\Dec}''),
           \emph{Optimization} (``\textup{\Opt}''), or \emph{Query answering} (``\textup{\Que}'') tasks.
           The remaining columns provide numbers of instances per empirical hardness class,
           distinguishing \emph{satisfiable} and \emph{unsatisfiable} instances 
           classified as \textup{\textbf{Easy}}, \textup{\textbf{Medium}}, or \textup{\textbf{Hard}},
           while \textup{\textbf{Too hard}} instances are divided into those known to be \emph{satisfiable}
           and others whose satisfiability is \emph{unknown}.
           For each hardness class and satisfiability status, the number in front of parentheses
           stands for the selected instances out of the respective available instances whose number is given
           in parentheses.\label{tab:benchmarks}}
 \end{table}
Table~\ref{tab:benchmarks} gives an overview of all problem domains,
grouped by their respective sub-tracks, of the Seventh ASP Competition,
where the names of new domains are highlighted in boldface.
The second column provides the computational task addressed in a domain,
distinguishing Decision (``\textup{\Dec}'') and Optimization
(``\textup{\Opt}'') problems as well as Query answering (``\textup{\Que}'').
Further columns categorize the instances in each domain by their empirical hardness,
where hardness classes are based on the performance
of the same reference systems, i.e., \clasp, \lptwoclaspold, and \waspdlvold,
as in the previous ASP Competition edition \cite{GebserMR17}:%
\footnote{This choice of reference systems allows us to reuse the runtime results
          for previous domains gathered in exhaustive experiments on all available
          instances that took about 212 CPU days on the competition platform.
          Instances that do not belong to any of the listed hardness classes are
          in the majority of cases ``very easy'' and the remaining ones
          ``non-groundable'', and we exclude such (uninformative regarding the
          system ranking) instances from the benchmark suite.}
\begin{itemize}
%
%
\item[$\bullet$] \textbf{Easy}:
Instances completed by at least one reference system in more than 20 seconds
and by all reference systems in less than 2 minutes solving time.
\item[$\bullet$] \textbf{Medium}:
Instances completed by at least one reference system in more than 2 minutes
and by all reference systems in less than 20 minutes (the competition timeout) solving time.
\item[$\bullet$] \textbf{Hard}:
Instances completed by at least one reference system in less than 40 minutes, 
while also at least one (not necessarily the same) reference system did not finish solving in 20 minutes.
\item[$\bullet$] \textbf{Too hard}:
Instances such that
none of the reference systems finished solving in 40 minutes.
\end{itemize}
For each of these hardness classes, numbers of available instances per problem domain
are shown within parentheses in Table~\ref{tab:benchmarks}, further distinguishing
satisfiable and unsatisfiable instances, whose respective numbers are given first or
second, respectively.
In case of instances classified as ``too hard'', however, no reference system could
report unsatisfiability, and thus the numbers of instances listed second refer to an
unknown satisfiability status.
Note that there are likewise no ``too hard'' instances of Decision problems or
Query answering domains known as satisfiable, so that the respective numbers are zero.
For example, the \Probl{Sokoban} domain features satisfiable as well as unsatisfiable
instances for each hardness class apart from the ``too hard'' one,
where 0 instances are known as satisfiable and 620 have an unknown satisfiability status.
Unlike that, ``too hard'' instances of Optimization problems are frequently known to
be satisfiable, in which case none of the reference systems was able to confirm an
optimum solution within 40 minutes.
Moreover, we discarded any instance of an Optimization problem that was reported to be
unsatisfiable, so that the respective numbers given second are zero for the first three
hardness classes.
This applies, e.g., to instances in the \Probl{Bayesian Network Learning} domain,
including 4, 8, 19, and 20 satisfiable instances that are ``easy'', ``medium'', ``hard'',
or ``too hard'', respectively, while the satisfiability status of further 6 ``too hard''
instances is unknown.
Finally, the numbers in front of parentheses in Table~\ref{tab:benchmarks} report how
many instances were (randomly) selected per hardness class and satisfiability status,
and the selection process is described in the rest of this section.

Given the numbers of satisfiable, unsatisfiable, or unknown in case of ``too hard''
instances per hardness class, our benchmark selection process aims at picking
20 instances in   each problem domain such that the four hardness classes are balanced,
while another share of instances is added freely.
Perfect balancing would then consist of picking four instances per hardness
class and another four instances freely in order to
guarantee that each hardness class contributes 20\% of the instances
in a domain.
Since in most domains the instances are not evenly distributed,
it is not possible though to insist on at least four instances per hardness class,
and rather we have to compensate for underpopulated classes at  which the number
of available instances is smaller.

The input to our balancing scheme includes a collection~$C$ of classes, where each
class is identified with the set of its contained instances.
The first step of balancing then determines the non-empty classes
from which instances can be picked:
\pagebreak[1]
\begin{eqnarray*}
 \mathit{classes} & \lesskip = \lesskip & \{x\in C \mid x\neq\emptyset\}.
\end{eqnarray*}
%
The number of non-empty classes is used to calculate how many instances should
ideally be picked per class, where the calculation makes use of the parameters
$n=20$ and $m=1$, standing for the total number of instances to select per domain
and the fraction of instances to pick freely, respectively:
\pagebreak[1]
\begin{eqnarray}\label{eq:target}
 \mathit{target} & \lesskip = \lesskip & \lfloor n/(|\mathit{classes}|+m)\rfloor.
\end{eqnarray}
%
To account for underpopulated classes, in the next step we calculate the
gap between the intended number of and the available instances in each class:
\pagebreak[1]
\begin{eqnarray*}
 \mathit{gap}(x) & \lesskip = \lesskip &
 \begin{cases}
   0 \text{ \ if \ } x\in C\setminus\mathit{classes} \\
   \mathit{target}-|x| \text{ \ if \ } x\in\mathit{classes}.
 \end{cases}
\end{eqnarray*}
%

\begin{example}\label{ex:selection:grace1}
In the \Probl{Graceful Graphs} domain, the ``easy'', ``medium'', ``hard'',
and ``too hard'' classes contain 3, 5, 30, or 21 instances, respectively,
when not (yet) distinguishing between satisfiable and unsatisfiable instances.
Since all four hardness classes are non-empty, 
we obtain
$\mathit{classes}=\{\text{``easy''},\linebreak[1]\text{``medium''},\linebreak[1]\text{``hard''},\linebreak[1]\text{``too hard''}\}$.
The calculation of instances to pick per class yields
$\mathit{target} = \lfloor 20/(4+1) \rfloor = 4$,
so that we aim at 4 instances per hardness class.
Along with the number of instances available in each class,
we then get
$\mathit{gap}(\text{``easy''})=4-3=1$,
$\mathit{gap}(\text{``medium''})=4-5=-1$,
$\mathit{gap}(\text{``hard''})=4-30=-26$, and
$\mathit{gap}(\text{``too hard''})=4-21=-17$.
Note that a positive number expresses underpopulation of a class relative to the
intended number of instances, while negative numbers indicate capacities to
compensate for such underpopulation.%
\eofex
\end{example}

Our next objective is to compensate for underpopulated classes
by increasing the number of instances to pick from other classes in a fair way.
Regarding hardness classes, our compensation scheme relies on the direct successor
relation $\prec$ given by
$\text{``easy''} \prec \text{``medium''}$,
$\text{``medium''} \prec \text{``hard''}$, and
$\text{``hard''} \prec \text{``too hard''}$.
We denote the strict total order obtained as the transitive closure of~$\prec$ by~$<$,
and its inverse relation by~$>$.
Moreover, we let $\circ$ below stand for either $<$ or~$>$ to specify calculations
that are performed symmetrically, such as determining the number of easier or harder
instances available to compensate for the (potential) underpopulation of a class:
\pagebreak[1]
\begin{eqnarray*}
 \mathit{available}(x)^\circ & \lesskip = \lesskip & \text{$\sum$}_{x'\circ x}\mathit{gap}(x').
\end{eqnarray*}
%
The possibility of compensation in favor of easier or harder instances is then
determined as follows:
\pagebreak[1]
\begin{eqnarray*}
 \mathit{compensate}(x)^\circ & \lesskip = \lesskip & \min\{(|\mathit{gap}(x)|+\mathit{gap}(x))/2,
                                                           (|\mathit{available}(x)^\circ|-\mathit{available}(x)^\circ)/2\}.
\end{eqnarray*}
The calculation is such that a positive gap, standing for the underpopulation of a class,
is a prerequisite for obtaining a non-zero outcome, and the availability of easier or
harder instances to compensate with is required in addition.
Given the compensation possibilities,
the following calculations decide about how many easier or harder instances, respectively,
are to be picked to resolve an underpopulation,
where the distribution should preferably be even and tie-breaking in favor of harder instances
is used as secondary criterion if the number of instances to compensate for is odd:%
\footnote{%
Given that $\mathit{distribute}(x)^>$ (or $\mathit{distribute}(x)^<$)
is limited by $\mathit{compensate}(x)^<$ (or $\mathit{compensate}(x)^>$),
the superscripts ``$^>$'' and ``$^<$'' refer to easier or harder instances, respectively, to 
be picked in addition.
This reading is chosen for a convenient notation in the specification of classes whose numbers of instances
are to be increased for compensation.}
\pagebreak[1]
\begin{eqnarray*}
 \mathit{distribute}(x)^> & \lesskip = \lesskip & 
                                \min\{\mathit{compensate}(x)^<,
                                      \max\{|\mathit{gap}(x)|-\mathit{compensate}(x)^>,
                                            \lfloor|\mathit{gap}(x)|/2\rfloor\}\}
\\
 \mathit{distribute}(x)^< & \lesskip = \lesskip &
                                \min\{\mathit{compensate}(x)^>,|\mathit{gap}(x)|-\mathit{distribute}(x)^>\}.
\end{eqnarray*}
%
%
It remains to choose classes whose numbers of instances are to be increased for compensation,
where we aim to distribute instances to closest classes with compensation capacities.
The following inductive calculation scheme accumulates instances to distribute
according to this objective:%
\pagebreak[1]
\begin{eqnarray*}
 \mathit{accumulate}(x)^\circ & \lesskip = \lesskip &
 \begin{cases} 
    0 \text{ \ if \ } \{x'\in C \mid x'\circ x\} = \emptyset \\
    \begin{array}{@{}r@{}}
    \mathit{accumulate}(x')^\circ+\mathit{distribute}(x')^\circ-\mathit{increase}(x')^\circ \text{ \ if \ } x'\circ x
\negspace
    \\
     \text{and \ } x'\prec x \text{ \ or \ } x\prec x'
    \end{array} 
 \end{cases}
\\
 \mathit{increase}(x)^< & \lesskip = \lesskip & \min\{\mathit{accumulate}(x)^<,(|\mathit{gap}(x)|-\mathit{gap}(x))/2\}
\\
 \mathit{increase}(x)^> & \lesskip = \lesskip & \min\{\mathit{accumulate}(x)^>,(|\mathit{gap}(x)|-\mathit{gap}(x))/2-\mathit{increase}(x)^<\}.
\end{eqnarray*}
In a nutshell, $\mathit{accumulate}(x)^<$ and $\mathit{accumulate}(x)^>$ express how many
easier or harder instances, respectively, ought to be distributed up to a class~$x$,
and $\mathit{increase}(x)^<$ and $\mathit{increase}(x)^>$ stand for corresponding increases
of the number of instances to be picked from~$x$.
The instances to increase with are then added to the original number of instances to pick from a class
as follows:%
\pagebreak[1]
\begin{eqnarray*}
 \mathit{select}(x) & \lesskip = \lesskip & |x|-(|\mathit{gap}(x)|-\mathit{gap}(x))/2+\mathit{increase}(x)^<+\mathit{increase}(x)^>.
\end{eqnarray*}
%

\begin{example}\label{ex:selection:grace2}
Given 
$\mathit{gap}(\text{``easy''})=1$,
$\mathit{gap}(\text{``medium''})=-1$,
$\mathit{gap}(\text{``hard''})=-26$, and
$\mathit{gap}(\text{``too hard''})=-17$
from Example~\ref{ex:selection:grace1}
for the \Probl{Graceful Graphs} domain,
we obtain the following numbers indicating the availability of easier instances:
$\mathit{available}(\text{``easy''})^<=\nolinebreak 0$,
$\mathit{available}(\text{``medium''})^<=1$,
$\mathit{available}(\text{``hard''})^<=1+(-1)=0$, and
$\mathit{available}(\text{``too hard''})^<=1+(-1)+(-26)=-26$.
Likewise, the available harder instances are expressed by
$\mathit{available}(\text{``too hard''})^>=\nolinebreak 0$,
$\mathit{available}(\text{``hard''})^>=-17$,
$\mathit{available}(\text{``medium''})^>=(-17)+(-26)=-43$, and
$\mathit{available}(\text{``easy''})^>=(-17)+(-26)+(-1)=-44$.
Again note that positive numbers like $\mathit{available}(\text{``medium''})^<=1$
represent a (cumulative) underpopulation, while negative numbers such as
$\mathit{available}(\text{``easy''})^>=-44$ indicate compensation capacities.

Considering ``easy'' instances, 
we further calculate
$\mathit{compensate}(\text{``easy''})^<=\min\{(|1|+1)/2,\linebreak[1](|0|-0)/2\}=0$ and
$\mathit{compensate}(\text{``easy''})^>=\min\{(|1|+1)/2,\linebreak[1](|{-44}|-(-44))/2\}=1$.
This tells us that we can     add one harder instance to compensate for the
underpopulation of the ``easy'' class, while
$\mathit{compensate}(x)^\circ=\mathit{compensate}(\text{``easy''})^<=0$ for the other
classes $x\in\{\text{``medium''},\linebreak[1]\text{``hard''},\linebreak[1]\text{``too hard''}\}$
and $\circ\in\{<,>\}$.
Given that instances to distribute are limited by compensation possibilities,
which are non-zero at underpopulated classes only,
it is sufficient to concentrate on ``easy'' instances in the \Probl{Graceful Graphs} domain.
This yields
$\mathit{distribute}(\text{``easy''})^>=\min\{0,\max\{1-1,\linebreak[1]0\}\}=0$ and
$\mathit{distribute}(\text{``easy''})^<=\min\{1,\linebreak[1]1-\nolinebreak 0\}=1$,
so that one harder instance is to be picked more.

The calculation of instance number increases to compensate for underpopulated classes then starts with
$\mathit{accumulate}(\text{``easy''})^<=0$,
$\mathit{increase}(\text{``easy''})^<=\min\{0,(|1|-1)/2\}=0$,
$\mathit{accumulate}(\text{``medium''})^<=0+1-0=1$,
$\mathit{increase}(\text{``medium''})^<=\min\{1,(|{-1}|-(-1))/2\}=1$, and
$\mathit{accumulate}(\text{``hard''})^<=1+0-1=0$.
That is, the instance to distribute from the underpopulated ``easy'' class
to some harder class increases the number of ``medium'' instances, while we obtain 
$\mathit{increase}(\text{``hard''})^<=\mathit{accumulate}(\text{``too hard''})^<=
 \mathit{increase}(\text{``too hard''})^<=0$
as well as 
$\mathit{accumulate}(x)^>=\mathit{increase}(x)^>=0$ for all
$x\in\{\text{``easy''},\linebreak[1]\text{``medium''},\linebreak[1]\text{``hard''},\linebreak[1]\text{``too hard''}\}$.
The final numbers of instances to pick per hardness class in the \Probl{Graceful Graphs} domain
are thus determined by
$\mathit{select}(\text{``easy''})=3-(|1|-1)/2+0+0=3$,
$\mathit{select}(\text{``medium''})=5-(|{-1}|-(-1))/2+1+0=5$,
$\mathit{select}(\text{``hard''})=30-(|{-26}|-(-26))/2+0+0=4$, and
$\mathit{select}(\text{``too hard''})=21-(|{-17}|-(-17))/2+0+0=4$.
Note that 16 instances are to be selected from particular hardness classes
in total,
sparing the four instances to be picked freely,
and also that our balancing scheme takes care of exchanging an ``easy'' for a
``medium'' instance.%
\eofex
\end{example}

After determining the numbers of instances to pick per hardness class,
we also aim to balance between satisfiable and unsatisfiable instances
within the same class.
In fact, the above balancing scheme is general enough to be reused for this
purpose by letting $C=\{\text{satisfiable}(x),\linebreak[1]\text{unsatisfiable}(x)\}$
consist of the subclasses of satisfiable or unsatisfiable instances, respectively,
in a hardness class $x$ that includes at least one instance known to be satisfiable or unsatisfiable.%
\footnote{Otherwise, all subclasses to pick instances from are empty,
                     which would lead to division by zero in \eqref{eq:target}.}
Moreover, the parameters $n$ and $m$ used in \eqref{eq:target} are fixed to
$n=\mathit{select}(x)$ and $m=0$, which reflect that the satisfiability status
should be balanced among all instances to be picked from $x$ without allocating
an additional share of instances to pick freely.
For the strict total order on the subclasses in~$C$, we use
$\text{satisfiable}(x) \prec \text{unsatisfiable}(x)$,
let $<$ denote the transitive closure of~$\prec$, and $>$
its inverse relation.

\begin{example}\label{ex:selection:grace3}
Reconsidering the \Probl{Graceful Graphs} domain, we obtain the following
number of instances to pick based on their satisfiability status:
$\mathit{select}(\linebreak[1]\text{satisfiable}(\linebreak[1]\text{``easy''}))=\nolinebreak 3$,
$\mathit{select}(\linebreak[1]\text{unsatisfiable}(\linebreak[1]\text{``easy''}))=\nolinebreak 0$,
$\mathit{select}(\linebreak[1]\text{satisfiable}(\linebreak[1]\text{``medium''}))=\nolinebreak 3$,
$\mathit{select}(\linebreak[1]\text{unsatisfiable}(\linebreak[1]\text{``medium''}))=\nolinebreak 1$,
$\mathit{select}(\linebreak[1]\text{satisfiable}(\linebreak[1]\text{``hard''}))=\nolinebreak 2$, and
$\mathit{select}(\linebreak[1]\text{unsatisfiable}(\linebreak[1]\text{``hard''}))=\nolinebreak 2$.
Note that
$\mathit{select}(\linebreak[1]\text{satisfiable}(x))+\mathit{select}(\linebreak[1]\text{unsatisfiable}(x))=\mathit{select}(x)$
for $x\in\{\text{``easy''},\linebreak[1]\text{``hard''}\}$,
while
$\mathit{select}(\linebreak[1]\text{satisfiable}(\linebreak[1]\text{``medium''}))+
 \mathit{select}(\linebreak[1]\text{unsatisfiable}(\linebreak[1]\text{``medium''}))=3+1=4<5=
 \mathit{select}(\text{``medium''})$.
The latter is due to rounding in $\mathit{target}=\lfloor 5/2 \rfloor = 2$, and then compensating
for the underpopulated unsatisfiable instances by increasing the number of satisfiable ``medium''
instances to pick by one.%
\eofex
\end{example}

For instances of Decision problems or Query answering domains,
we have that secondary balancing based on the satisfiability status is
generally void for ``too hard'' instances, of which none are known
to be satisfiable or unsatisfiable.
In case of Optimization problems, where we discard instances known as unsatisfiable,
$\mathit{select}(\text{satisfiable}(x))=\mathit{select}(x)$ holds for
$x\in\{\text{``easy''},\linebreak[1]\text{``medium''},\linebreak[1]\text{``hard''}\}$,
while our balancing scheme favors ``too hard'' instances known as satisfiable
over those with an unknown satisfiability status.
This approach makes sure that ``too hard'' instances to be picked possess solutions, yet confirming
an optimum is hard, and instances with an unknown satisfiability status can still be
contained among those that are picked freely.

\begin{example}\label{ex:selection:valves}
Regarding the Optimization problem in the \Probl{Valves Location} domain, we obtain
$\mathit{select}(\linebreak[1]\text{satisfiable}(\linebreak[1]\text{``too hard''}))=\mathit{select}(\linebreak[1]\text{``too hard''})=4$,
given that the 23 instances whose satisfiability status is unknown are not considered for balancing.%
\eofex
\end{example}

The described twofold balancing scheme, first considering the hardness of instances and then
the satisfiability status of instances of similar hardness,
is implemented by an \aspcoretwo encoding that consists of two parts:
a deterministic program part (having a unique answer set)
takes care of determining the numbers $\mathit{select}(x)$ from the
runtime results of reference systems,
and a non-deterministic part similar to the selection program
used in the previous ASP Competition edition \cite{GebserMR17}
encodes the choice of 20 instances per domain such that lower bounds
given by the calculated numbers $\mathit{select}(x)$ are met.
In comparison to the previous competition edition, we updated the deterministic part of
the benchmark selection encoding by implementing the balancing scheme described above,
which is more general than before and not fixed to a particular number of classes
(regarding hardness or satisfiability status) to balance.
The instance selection was then performed by running the ASP solver \clasp\
with the options \texttt{--rand-freq}, \texttt{--sign-def}, and \texttt{--seed} for guaranteeing
reproducible randomization,
using the concatenation of winning numbers in the EuroMillions lottery of 2nd May 2017
as the random seed.
This process led to the numbers of instances picked per domain, hardness class, and satisfiability
status listed in Table~\ref{tab:benchmarks}.

As a final remark, we note that we had to exclude the \Probl{Resource Allocation} domain from the main
competition in view of an insufficient number of instances belonging to the hardness classes
under consideration.
In fact, the majority of instances turned out to be ``very easy'' relative to an optimized encoding
devised in the phase of checking/establishing the \aspcoretwo\ compliance of initial
submissions by benchmark authors.
This does not mean that the problem of \Probl{Resource Allocation} as such would be trivial or
uninteresting, but rather time constraints on running the main competition did unfortunately not permit
to extend and then reassess the collection of instances.


%% file: participants.tex
\section{Participant Systems}\label{sec:participants}

Fourteen systems, registered by three teams, participate in the
System track of the Seventh ASP Competition.
The majority of 
systems runs in the {\bf SP} category,
while 
two (indicated by the suffix ``{\sc {\small -mt}}'' below)
exploit parallelism in the {\bf MP} category.
In the following, we survey the registered teams and systems.

\paragraph{Aalto.}
    The team from Aalto University submitted nine systems that 
    utilize normalization \cite{bogeja14a,bogeja16a} and translation
    \cite{bojata16a,bogejakasc16a,gebs-etal-2014,JN11:scmcs,ljn-kr-12} means. 
    Two systems,
    {\sc {\small lp2sat+lingeling}} and {\sc {\small lp2sat+plingeling-mt}},
    perform translation to SAT and use
    {\sc {\small lingeling}} or {\sc {\small plingeling}}, respectively,
    as back-end solver.
    Similarly,
    {\sc {\small lp2mip}} and {\sc {\small lp2mip-mt}}
    rely on translation to Mixed Integer Programming along with a single-
    or multi-threaded variant of {\sc {\small cplex}} for solving.
    The {\sc {\small lp2acycasp}}, {\sc {\small lp2acycpb}}, and {\sc {\small lp2acycsat}} systems
    incorporate translations based on acyclicity checking,
    supported by {\sc {\small clasp}} run as ASP, Pseudo-Boolean, or SAT solver,
    as well as the {\sc {\small graphsat}} solver in case of SAT with acyclicity checking.
    Moreover, {\sc {\small lp2normal+lp2sts}} takes advantage of the {\sc {\small sat-to-sat}}
    framework to decompose complex computations into several SAT solving tasks.
    Unlike that, {\sc {\small lp2normal+clasp}} confines preprocessing to the (selective) normalization
    of aggregates and weak constraints before running {\sc {\small clasp}} as ASP solver.
    Beyond syntactic differences between target formalisms,
    the main particularities of the available translations concern space complexity and the
    supported language features.
    Regarding space, the translation to SAT utilized by
    {\sc {\small lp2sat+lingeling}} and {\sc {\small lp2sat+plingeling-mt}}
    comes along with a logarithmic overhead in case of non-tight logic programs
    that involve positive recursion 
    \cite{fage-94},
    while the other translations are guaranteed to remain linear.
    Considering language features,
    the systems by the Aalto team do not support queries,
    and the back-end solver {\sc {\small clasp}} of 
    {\sc {\small lp2acycasp}}, {\sc {\small lp2acycpb}}, and {\sc {\small lp2normal+clasp}}
    provides a native implementation of aggregates,
    which the other systems treat by normalization within preprocessing.
    Optimization problems are supported by all systems but
    {\sc {\small lp2sat+lingeling}}, {\sc {\small lp2sat+plingeling-mt}}, and
    {\sc {\small lp2normal+lp2sts}},
    while only {\sc {\small lp2normal+lp2sts}} and {\sc {\small lp2normal+clasp}}
    are capable of handling non-HCF disjunction.

\paragraph{ME-ASP.}
The ME-ASP team from the University of Genova, the University of Sassari, and the University of Calabria
submitted the multi-engine ASP system {\sc {\small me-asp2}},
which is an updated version of {\sc {\small me-asp}} \cite{mara-etal-jeli-2012,mara-etal-2014-tplp,mara-etal-2015-lpnmr},
the winner system in the Regular track of the Sixth ASP Competition.
Like its predecessor version,
{\sc {\small me-asp2}} investigates features of an input program to select its back-end
among a pool of ASP grounders and solvers.
Basically, {\sc {\small me-asp2}} applies algorithm selection techniques before each stage of the answer set computation, with the goal of selecting the most promising computation strategy overall.
As regards grounders, {\sc {\small me-asp2}} can pick either {\sc {\small dlv}} or \Gringo,
while the available solvers include a selection of those submitted to the Sixth ASP Competition 
as well as the latest version of {\sc {\small clasp}}. 
The first selection (basically corresponding to the selection of the grounder) is based on features of non-ground programs and was obtained by implementing the result of the application of the PART decision list algorithm, whereas the choice of a solver is based on the multinomial classification algorithm k-Nearest Neighbors, used to train a model on features of ground programs extracted (whenever required) from the output generated by the grounder (for more details, see \cite{mara-etal-2015-lpnmr}).




\paragraph{UNICAL.}
The team from the University of Calabria submitted four systems
utilizing the recent {\sc {\small idlv}} grounder \cite{DBLP:journals/ia/CalimeriFPZ17},
developed as a redesign of (the grounder component of) {\sc {\small dlv}} going along with the addition of new features.
Moreover, back-ends for solving are selected from a variety of existing ASP solvers. 
In particular, {\sc {\small idlv-clasp-dlv}} makes use of {\sc {\small dlv}} \cite{DBLP:journals/tocl/LeonePFEGPS06,mara-etal-2008-joacil} 
for instances featuring a ground query; otherwise, it consists of the combination of the grounder {\sc {\small idlv}} with {\sc {\small clasp}} 
executed with the option \texttt{--configuration=trendy}. 
The {\sc {\small idlv+-clasp-dlv}} system is a variant of the previous system that uses a heuristic-guided rewriting technique \cite{DBLP:conf/padl/CalimeriFPZ18} relying on hyper-tree decomposition, which aims to automatically replace long rules with sets of smaller ones that are possibly evaluated more efficiently. 
{\sc {\small idlv+-wasp-dlv}} is obtained by using {\sc {\small wasp}} 
in place of {\sc {\small clasp}}. In more detail, {\sc {\small wasp}} is executed with the options \texttt{--shrinking-strategy=progression} \texttt{--shrinking-budget=10 --trim-core --enable-disjcores}, which configure {\sc {\small wasp}} to use two techniques tailored for Optimization problems.
Inspired by {\sc {\small me-asp}}, 
{\sc {\small idlv+s}} \cite{DBLP:conf/aiia/FuscaCZP17} integrates {\sc {\small idlv+}} with an automatic selector to choose between {\sc {\small wasp}} and {\sc {\small clasp}} on a per instance basis. To this end, {\sc {\small idlv+s}} implements  classification, by means of the well-known support vector machine technique. A more detailed description of the {\sc {\small idlv+s}} system is provided in \cite{DBLP:journals/corr/abs-1810-00041}.

%% file: results.tex
\section{Results}
\label{sec:results}

This section presents the results of the Seventh ASP Competition. We first announce the winners in the {\bf SP} category and analyze their performance, and then proceed overviewing results in the {\bf MP} category. Finally, we analyze the results more in details outlining some of the outcomes.

\subsection{Results in the {\bf SP} Category}

\begin{figure}[t]
\begin{center}
\includegraphics[width=\textwidth]{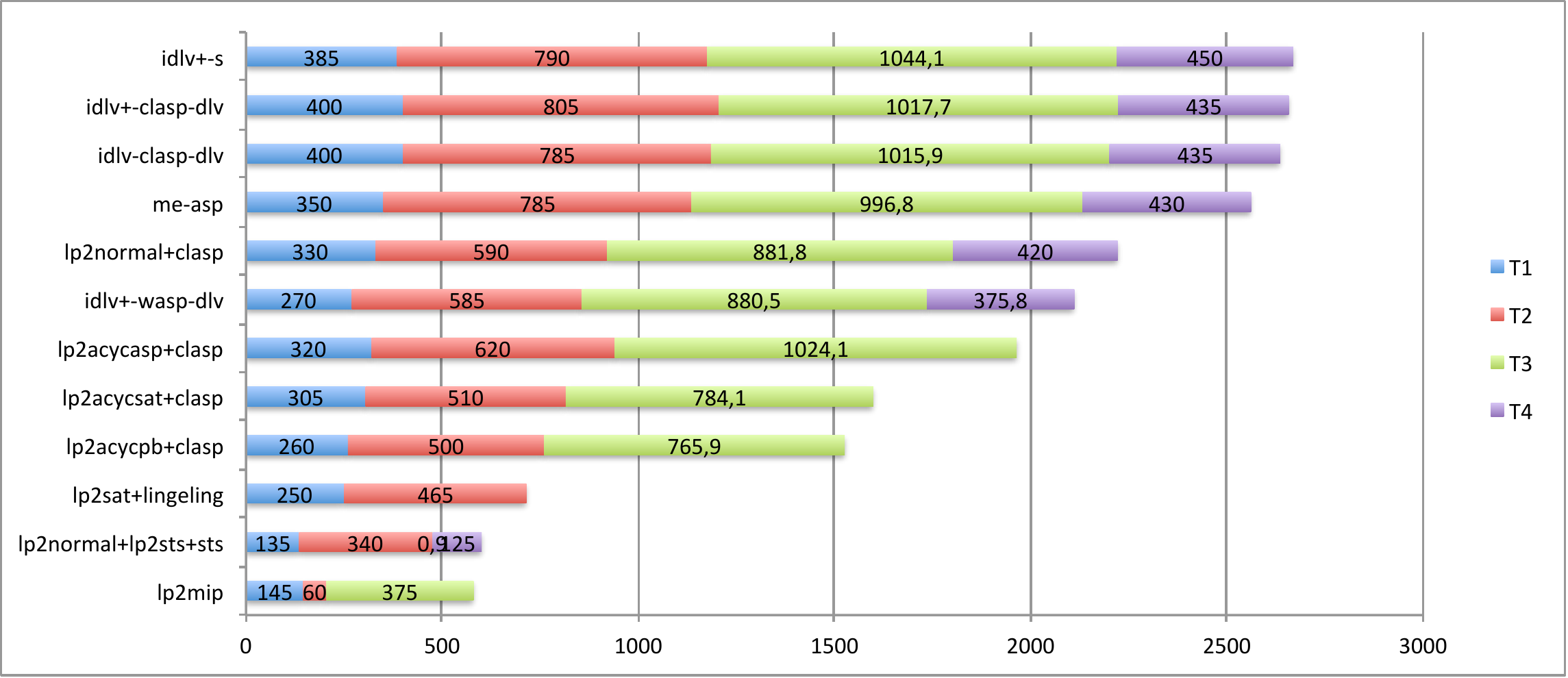}
\end{center}
\caption{Results of the {\bf SP} category with computation $S_1$ for Optimization problems.} 
\label{fig:SP-S_1}
\end{figure}

\begin{figure}[t]
\begin{center}
\includegraphics[width=\textwidth]{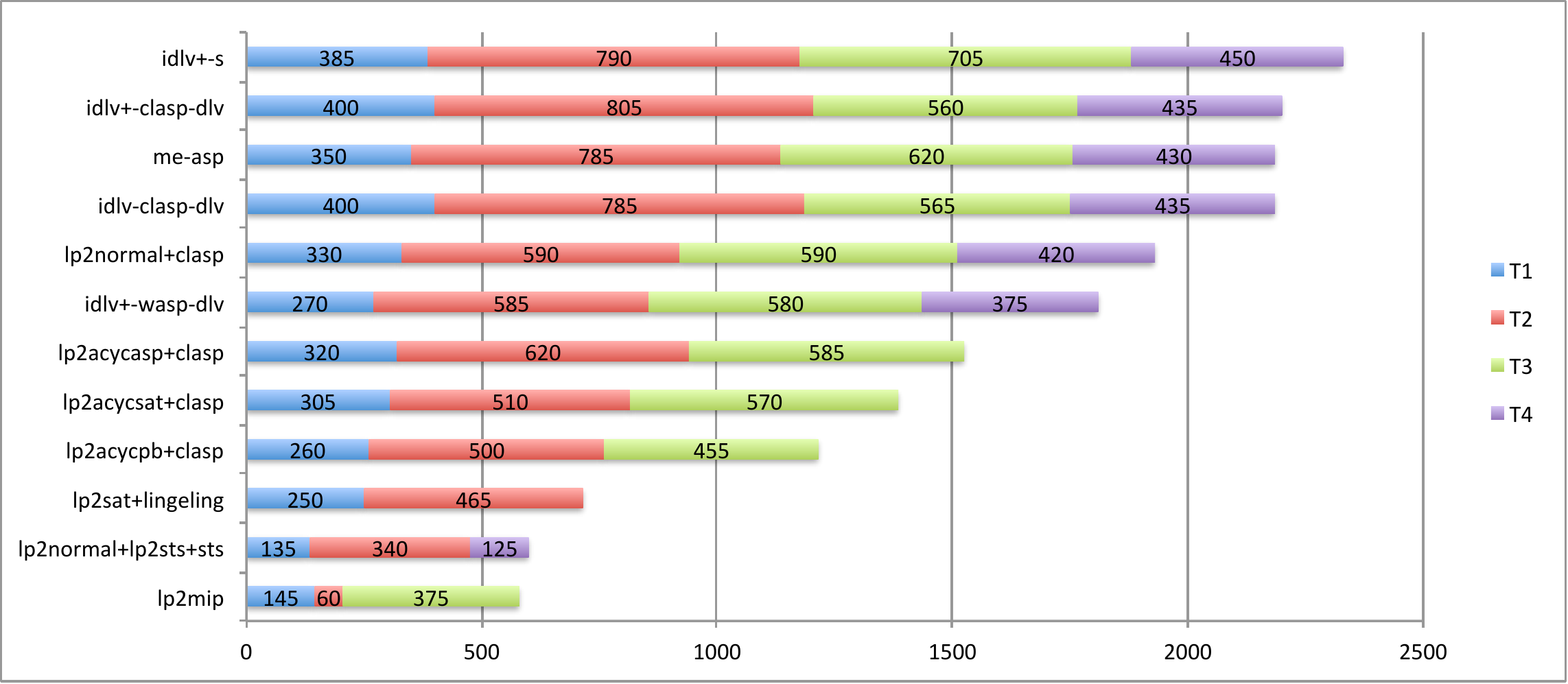}
\end{center}
\caption{Results of the {\bf SP} category with computation $S_2$ for Optimization problems.} 
\label{fig:SP-S_2}
\end{figure}

Figures~\ref{fig:SP-S_1} and ~\ref{fig:SP-S_2} summarize the results of the {\bf SP} category, by showing the scores of the various systems, where Figure~\ref{fig:SP-S_1} utilizes function $S_1$ for computing the score of Optimization problems, while Figure~\ref{fig:SP-S_2} utilizes function $S_2$. 
To sum up, considering Figure~\ref{fig:SP-S_1}, the first three places go to the systems:
\begin{enumerate}
\item {\sc idlv+s}, by the UNICAL team, with 2665 points; 
\item {\sc idlv+-clasp-dlv}, by the UNICAL team, with 2655,9 points; 
\item {\sc idlv-clasp-dlv}, by the UNICAL team, with 2634 points. 
\end{enumerate}

Also, {\sc me-asp} is quite close in performance, earning 2560 points in total.

Thus, the first three places are taken by versions of the {\sc idlv} system pursuing the approaches outlined in Section~\ref{sec:participants}. The fourth place, with very positive results, is instead taken by {\sc me-asp} which pursues a portfolio approach, and was the winner of the last competition. 

Going into the details of sub-tracks, 
the three top-performing systems overall take the first places as well:

 \begin{itemize}
 \item Sub-track \#1 (Basic Decision):    {\sc idlv-clasp-dlv} and {\sc idlv+-clasp-dlv} with 400 points;
 \item Sub-track \#2 (Advanced Decision): {\sc idlv+-clasp-dlv}     with 805 points;
 \item Sub-track \#3 (Optimization):      {\sc idlv+-clasp-dlv} with 1015,9 points;
 \item Sub-track \#4 (Unrestricted):      {\sc idlv+s}    with 450 points.
 \end{itemize}

Considering Figure~\ref{fig:SP-S_2}, which employs function $S_2$ for computing scores of Optimization problems, the situation is slightly different, i.e., {\sc me-asp} now gets the third place. To sum up, the first three places go to the systems:

\begin{enumerate}
\item {\sc idlv+s}, by the UNICAL team, with 2330 points; 
\item {\sc idlv+-clasp-dlv}, by the UNICAL team, with 2200 points; 
\item {\sc me-asp}, by the ME-ASP team, with 2185 points. 
\end{enumerate}

{\sc idlv+-clasp-dlv} is now fourth with the same score of 2185 points but higher cumulative CPU time: the difference is in Sub-track \#3, where relative results are different with respect to using $S_1$, and with the new score computation {\sc me-asp} earns 55 points more than {\sc idlv+-clasp-dlv}. In general, employing $S_2$ function for computing scores of Optimization problems leads to lower scores: indeed, $S_2$ is more restrictive than $S_1$ given that only optimal results are considered.

An overall view of the performances of all participant systems on all benchmarks is shown in the cactus plot of Figure~\ref{fig:cactus}. Detailed results are reported in Appendix A.

\begin{figure}[t]
\begin{center}
\includegraphics[width=\textwidth]{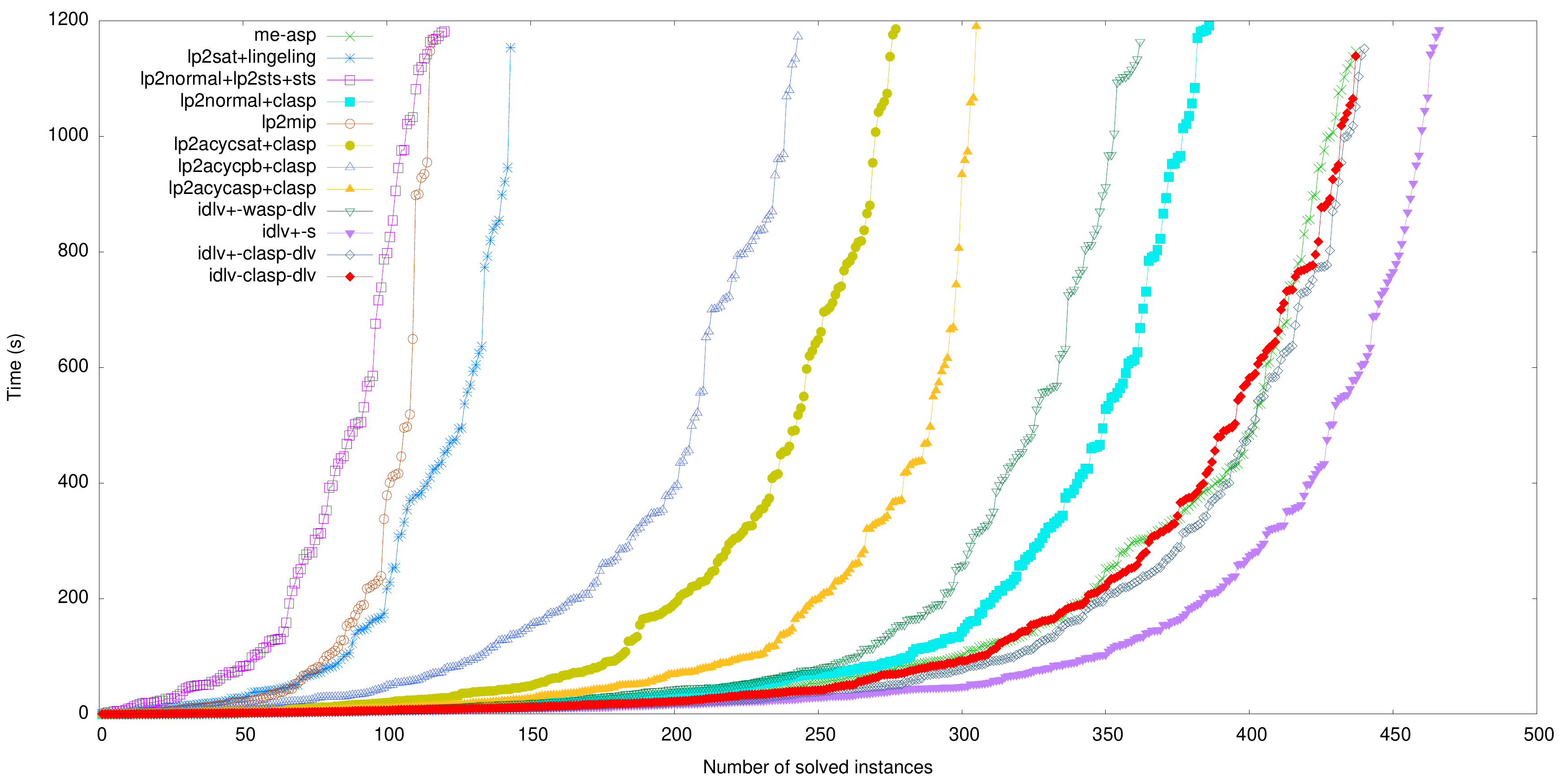}
\end{center}
\caption{Cactus plot of solver performances in the {\bf SP} category.} 
\label{fig:cactus}
\end{figure}

\begin{figure}[t]
\begin{center}
\includegraphics[width=\textwidth]{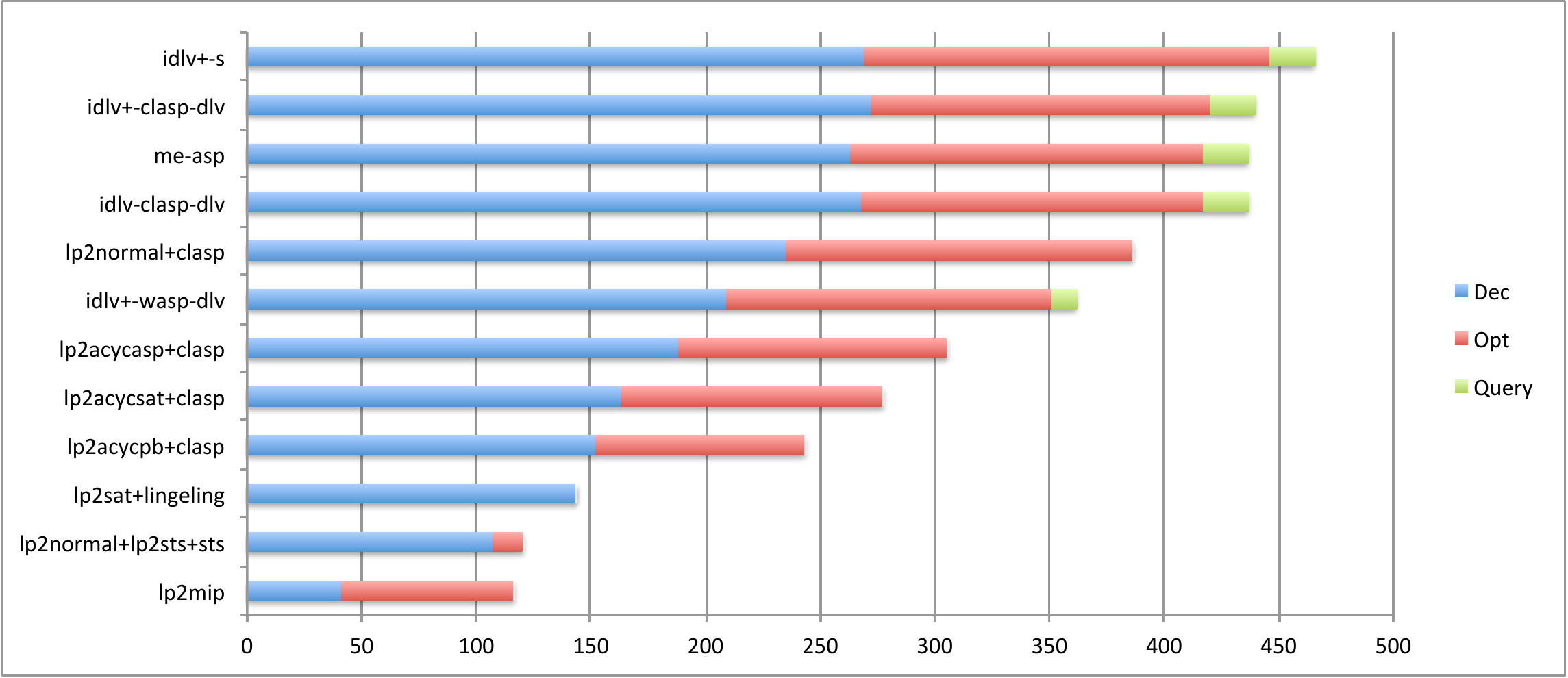}
\end{center}
\caption{Number of (optimally) solved instances in the {\bf SP} category by task.} 
\label{fig:SolvedByTask}
\end{figure}

\begin{figure}[t]
\begin{center}
\includegraphics[width=\textwidth]{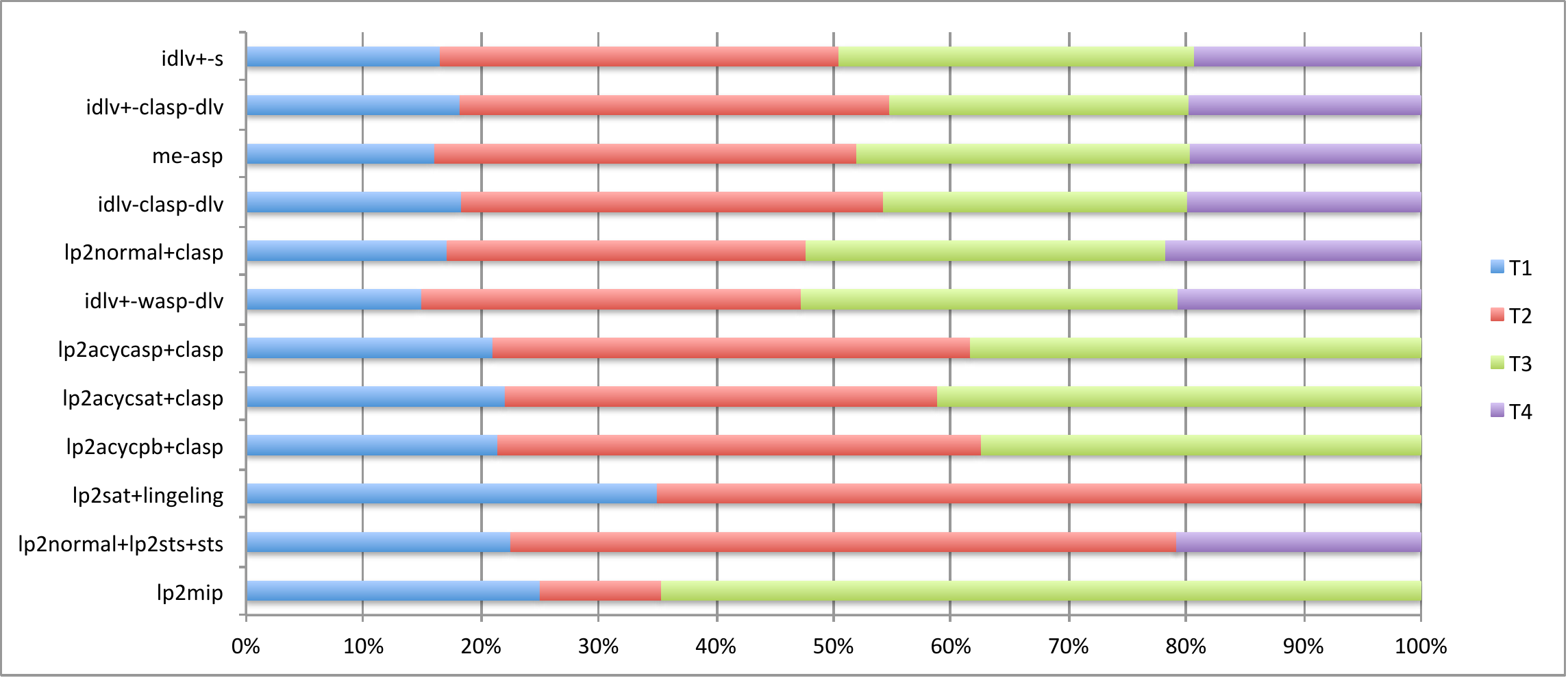}
\end{center}
\caption{Percentage of solved instances in the {\bf SP} category.} 
\label{fig:PercentageSolved}
\end{figure}

Official results in Figures~\ref{fig:SP-S_1} and~\ref{fig:SP-S_2} are complemented by the data showed in Figures~\ref{fig:SolvedByTask} and~\ref{fig:PercentageSolved}. Figure~\ref{fig:SolvedByTask} contains, for each solver, the number of (optimally) solved instances in each reasoning problem of the competition, i.e., Decision, Optimization and Query (denoted Dec, Opt, and Query in the figure, respectively). From the figure, we can see that {\sc idlv+-clasp-dlv} and {\sc idlv-clasp-dlv} are the solvers that perform best on Decision problems, while {\sc idlv+-s} is the best on Optimization problems. For what concern Query answering, the first four solvers perform equally well on them. Figure~\ref{fig:PercentageSolved}, instead, reports, for each solver, the percentage of solved instances (resp. score) in the various sub-tracks out of the total number of (optimally) solved instances (resp. global score), i.e., what is the ``contribution'' of tasks in each sub-track to the results of the respective system.

\begin{figure}[t]
\begin{center}
\includegraphics[width=\textwidth]{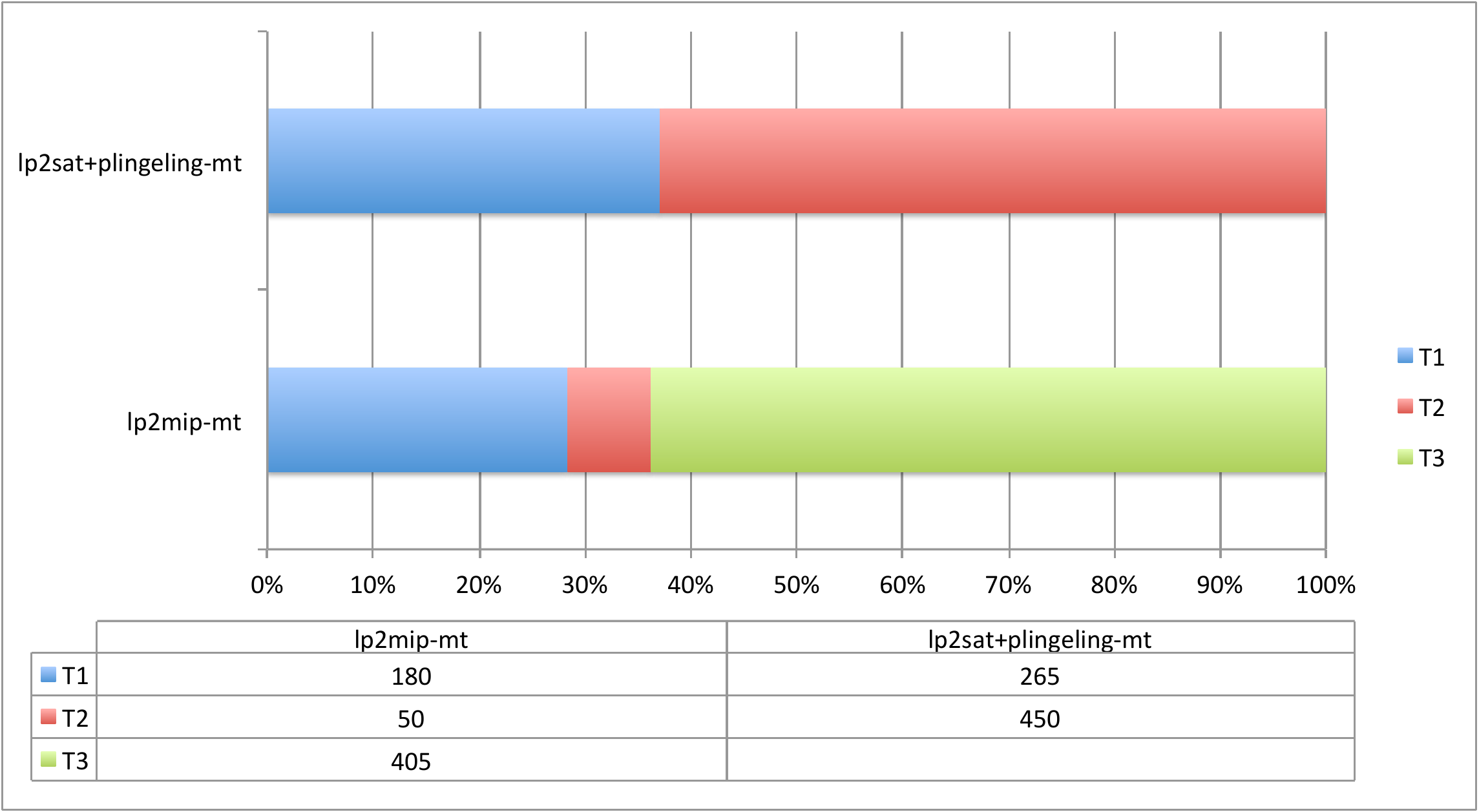}
\end{center}
\caption{Results of the {\bf MP} category.} 
\label{fig:MP}
\end{figure}

\subsection{Results in the {\bf MP} category}

Figure~\ref{fig:MP} shows results about the {\bf MP} category. The bottom part of the figure reports the scores acquired by the two participant systems, which cumulatively are: for

\begin{enumerate}
\item {\sc lp2sat+plingeling-mt}, by the Aalto team, 715 points; 
\item {\sc lp2mip-mt}, by the Aalto team, 635 points.
\end{enumerate}

Looking into details of the sub-tracks, we can note that {\sc lp2sat+plingeling-mt} is better than {\sc lp2mip-mt} on Sub-track \#1 and much better on Sub-track \#2, while on Sub-track \#3, where {\sc lp2sat+plingeling-mt} does not compete, {\sc lp2mip-mt} earns a consistent number of points, but not enough to globally reach the score of {\sc lp2sat+plingeling-mt} in the first two sub-tracks.

The top part of Figure~\ref{fig:MP}, instead, complements the results by showing the ``contribution'' of solved instances in each sub-track out of the score of the respective system.

\subsection{Analysis of the results} There are some general observations that can be drawn out of the results presented in this section. First, the best overall solver implements algorithm selection techniques, and continues the ``tradition'' of the efficiency of portfolio-based solvers in ASP competitions, given that {\sc \small claspfolio}~\cite{gebs-etal-2011-claspfolio} and {\sc \small me-asp}~\cite{mara-etal-2014-tplp} were the overall winners of the 2011 and 2015 competitions, respectively.
At the same time, the result outlines the importance of the introduction of new evaluation techniques and implementations. Indeed, although {\sc idlv+-s} applies a strategy similar to the one of {\sc \small me-asp}, {\sc idlv+-s} exploited a new component (i.e., the grounder~\cite{cali-etal-aiia-2016}) that was not present in  {\sc \small me-asp} (which is based on solvers from the previous competition).
Second, the approach implemented by {\sc {\small lp2normal}} using {\sc clasp} confirms its very good behavior in all sub-tracks, and thus overall. Third, specific istantiations of the translation-based approach perform particularly well in some sub-tracks: this is the case for the {\sc {\small lp2acycasp}} solver using {\sc\small clasp} in Sub-track \#3, especially when considering scoring scheme $S_1$, but also for {\sc {\small lp2mip}}, that compiles to a general purpose solver, in the same sub-track, especially when considering scoring scheme $S_2$ (even if to a less extent). As far as the comparison between solvers in the {\bf MP} category and their counter-part in the {\bf SP} category is concerned, we can see that globally the score of {\sc lp2sat+plingeling-mt} and {\sc lp2sat+plingeling} is the same, with the small advantage of {\sc lp2sat+plingeling-mt} in Sub-track \#1 being compensated in Sub-track \#2. Instead, {\sc {\small lp2mip+mt}} earns a consistent number of points more than {\sc {\small lp2mip}}, especially in Sub-track \#1 and \#3. In general, more specific research is probably needed on ASP solvers exploiting multi-threading to take real advantage from this setting.


%% file: conclusions.tex
\section{Conclusion and Final Remarks}
\label{sec:conclusions}

We have presented design and results of the Seventh ASP Competition,
with particular focus on new problem domains, revised benchmark selection process, systems registered for the event, and results. 


In the following, we draw some recommendations for future editions. These resemble the ones of the past event: for some of them some steps have been already made in this seventh's event, but they may be considered, with the aim of widening the number of participant systems and application domains that can be analyzed, starting from the next (Eighth) ASP competition that will take place in 2019 in affiliation with the 15th International Conference on Logic Programming and Non-Monotonic Reasoning (LPNMR 2019), in Philadelphia, US:

\begin{itemize}
\item We also tried to re-introduce a Model\&Solve track at the competition. But, given the short call for contributions and the low number of expressions of interest received, we decided not to run the track. Despite this, we still think that a (restricted form of a) Model\&Solve track should be re-introduced in the ASP competition series.
\item
Our aim with the re-introduction of a Model\&Solve track was at solving domains involving, e.g.,
discrete as well as continuous dynamics~\cite{BalducciniMML17},
so that extensions like
Constraint Answer Set Programming~\cite{mell-etal-amai-2008}
and incremental ASP solving~\cite{gekakaosscth08a}
may be exploited. The mentioned extensions could be added as tracks of the competition, but for CASP the first step that would be needed is a standardization of its language.
\item Given that still basically all participant systems rely on grounding, the availability of more grounders is crucial. In this event the {\sc \small I-DLV} grounder came into play, but there is also the need for more diverse techniques. This may also help improving portfolio solvers, by exploiting machine learning techniques at non-ground level (for a preliminary investigation, see~\cite{mara-etal-2013-aiia,mara-etal-2015-lpnmr}).

\item Portfolio solvers showed good performance in the editions where they participated. However, no such system in the various editions has exploited
a parallel portfolio approach. Exploring such techniques in conjunction could be an interesting topic of future research for further improving the efficiency. 
\item Another option for attracting (young) researchers from neighboring areas to the development of ASP solvers may be a track dedicated to modifications of a common reference system, in the spirit of the Minisat hack track of the SAT Competition series. This would lower the “entrance barrier” by keeping the effort of a participation affordable, even for small teams.

\end{itemize}

%% file: paper.bbl
\newcommand{\SortNoOp}[1]{}

%% file: appendix.tex
\appendix
\section{Detailed Results}

\begin{figure}[ht!]
\begin{center} 
\includegraphics[trim={10 260 30 10},width=.9\textheight,angle=90]{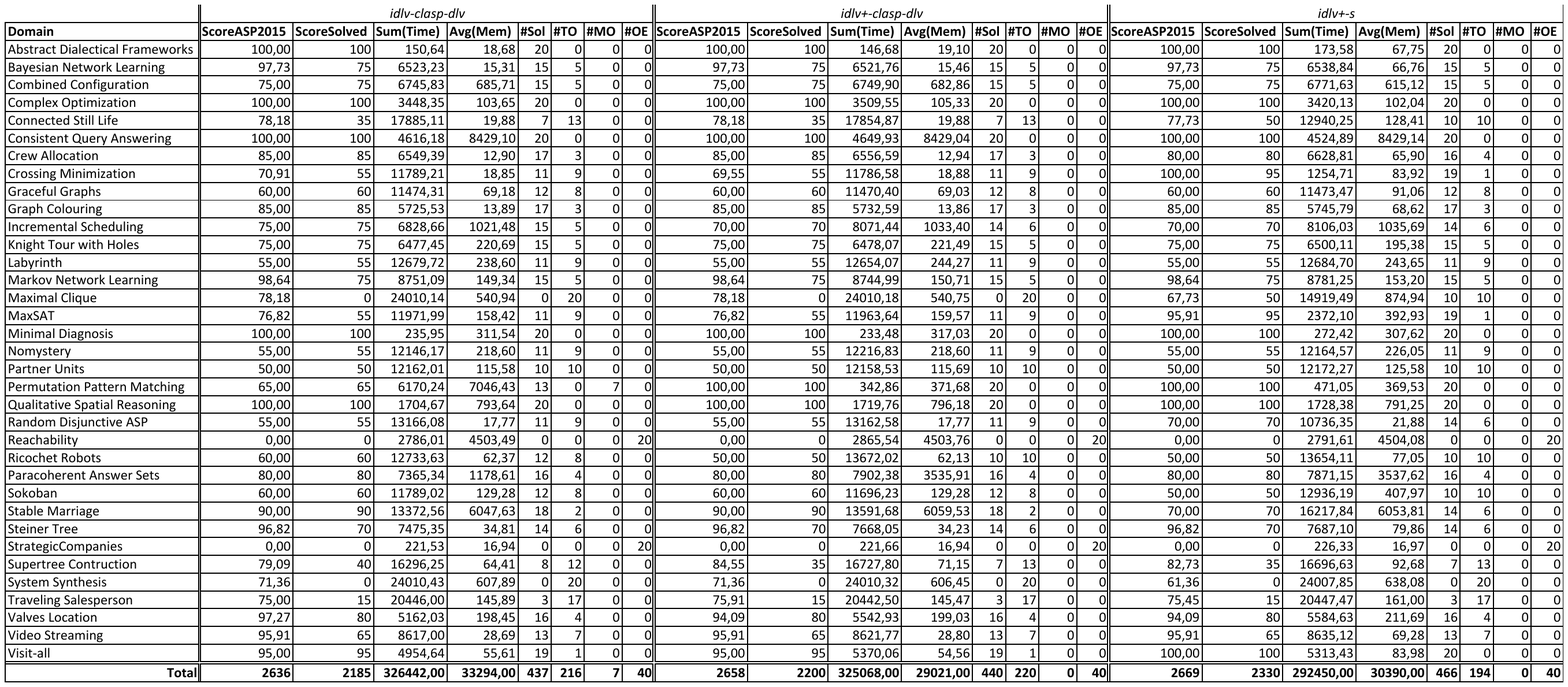}
\end{center}
\caption{Detailed results for {\sc idlv-clasp-dlv}, {\sc idlv+-clasp-dlv}, {\sc idlv+-s}.} 
\label{fig:bigtable:1}
\end{figure}

\begin{figure}[ht!]
\begin{center} 
\includegraphics[trim={10 260 30 10},width=.9\textheight,angle=90]{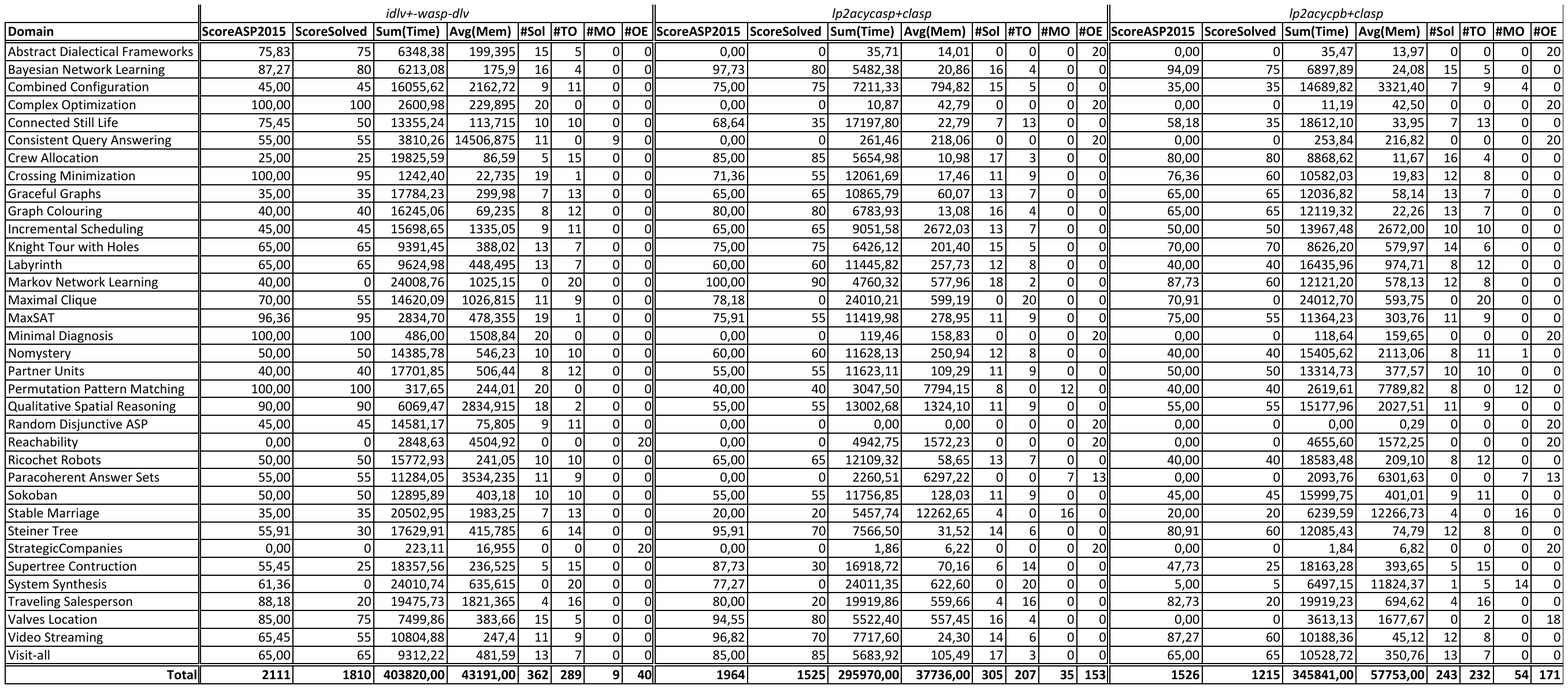}
\end{center}
\caption{Detailed results for {\sc idlv-wasp-dlv}, {\sc lp2acycasp+clasp}, {\sc lp2acycpb+clasp}.} 
\label{fig:bigtable:2}
\end{figure}

\begin{figure}[ht!]
\begin{center} 
\includegraphics[trim={10 260 30 10},width=.9\textheight,angle=90]{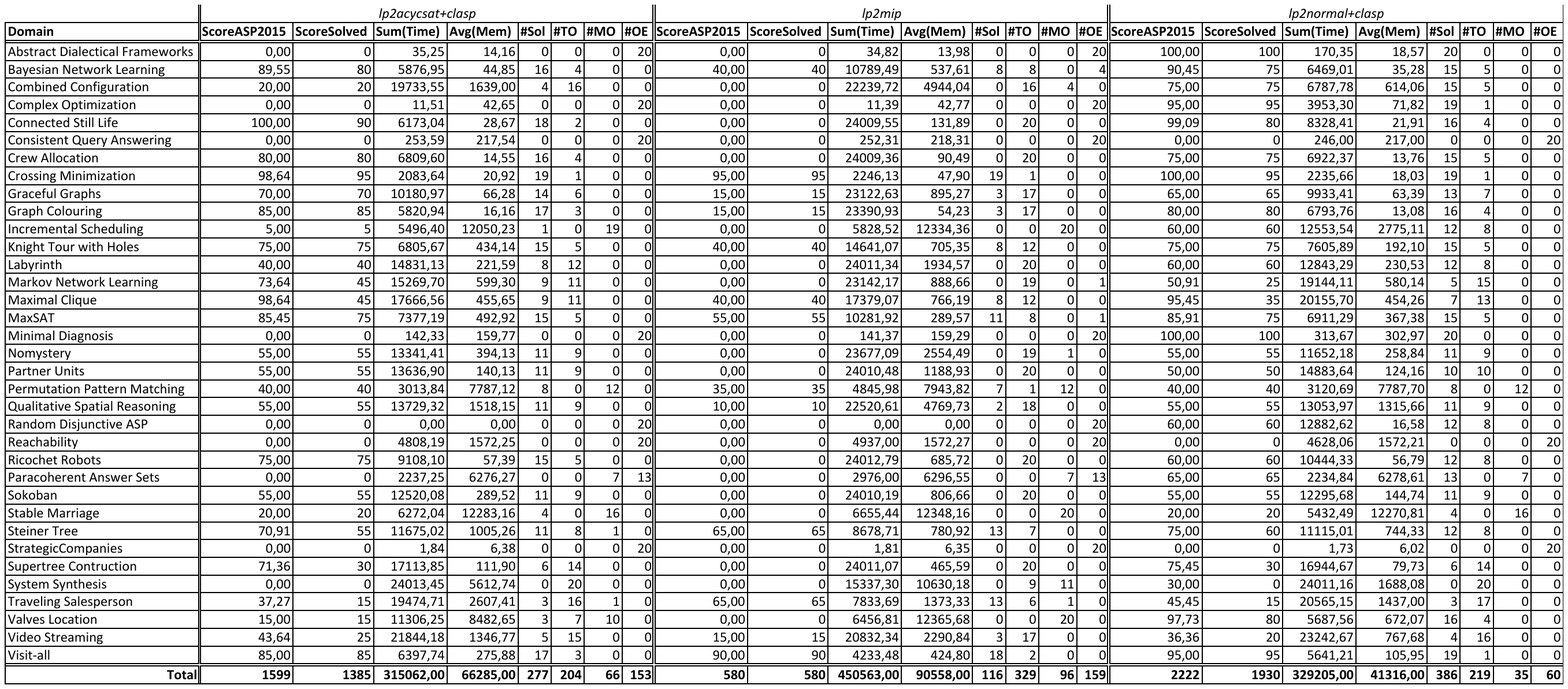}
\end{center}
\caption{Detailed results for {\sc lp2acycsat+clasp}, {\sc lp2mip}, {\sc lp2normal+clasp}.} 
\label{fig:bigtable:3}
\end{figure}

\begin{figure}[ht!]
\begin{center} 
\includegraphics[trim={10 260 30 10},width=.9\textheight,angle=90]{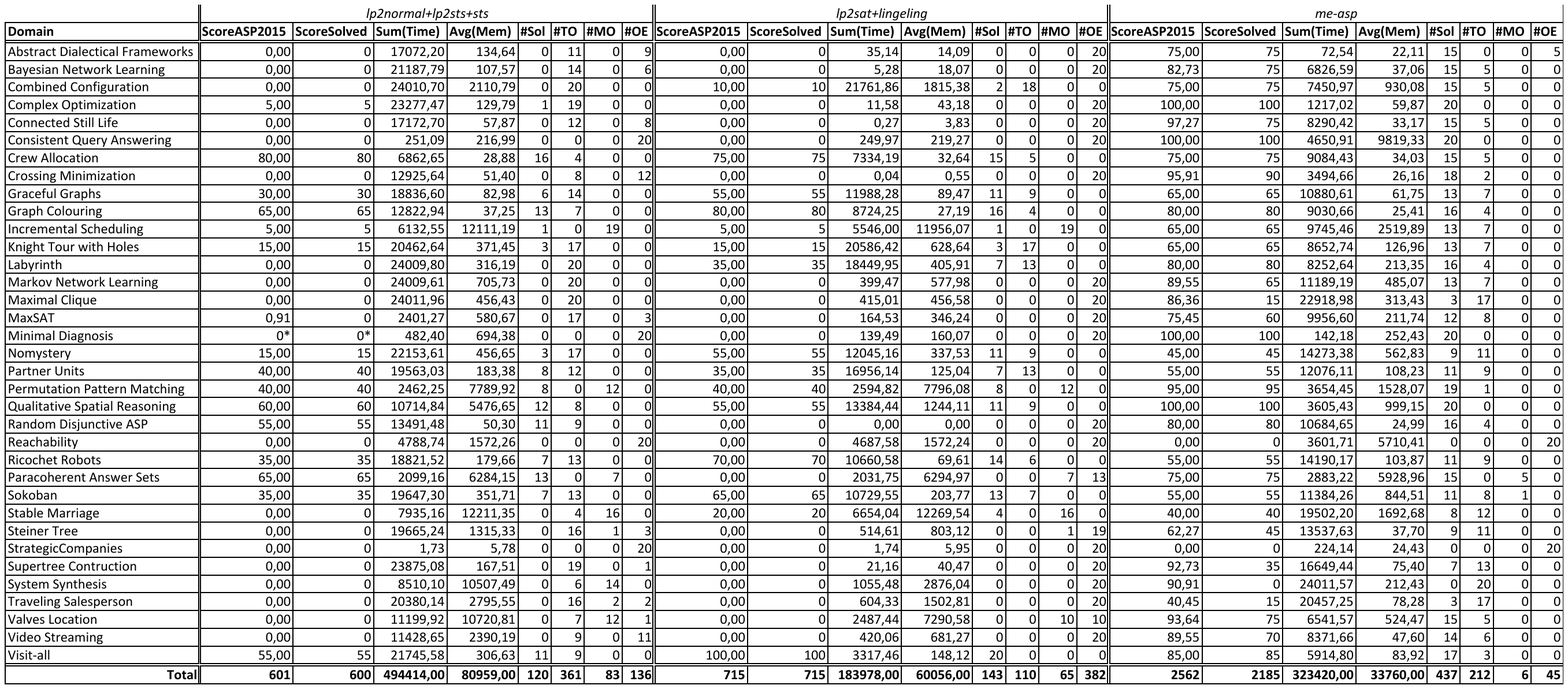}
\end{center}
\caption{Detailed results for {\sc lp2normal+lp2sts+sts}, {\sc lp2sat+lingeling}, {\sc me-asp}.} 
\label{fig:bigtable:4}
\end{figure}

We report in this appendix the detailed results aggregated by solver.
In particular, Figures~\ref{fig:bigtable:1}-\ref{fig:bigtable:4} report for each solver and for each domain the score with computation $S_1$ for Optimization problems (ScoreASP2015), with computation score $S_2$ for Optimization problems (ScoreSolved), the sum of the execution times for all instances (Sum(Time)), the average memory usage on solved instances (Avg(Mem)), the number of solved instances (\#Sol), the number of timed out executions (\#TO), the number of execution terminated because the solver exceeded the memory limit (\#MO) and the number of execution with abnormal execution (\#OE), this last counting the instances that could not be solved by a solver, thus including output errors, abnormal terminations, give-ups as well as instances that cannot be solved by a solver when it did not participate to a domain. An ``*'' near to a score of 0 indicates that the solver was disqualified from a domain because it terminated normally but produced a wrong witness in some instance of the domain.

\newpage
\newpage
\newpage
\newpage

